\documentclass{article}

\PassOptionsToPackage{numbers, compress}{natbib}
\usepackage[final]{neurips_2021}

\usepackage[utf8]{inputenc} %
\usepackage[T1]{fontenc}    %
\usepackage[colorlinks=true,citecolor=black,urlcolor=black,linkcolor=black]{hyperref}       %
\usepackage{url}            %
\usepackage{booktabs}       %
\usepackage{amsfonts}       %
\usepackage{nicefrac}       %
\usepackage{microtype}      %
\usepackage{xcolor}         %

\usepackage{amsmath}
\usepackage[]{hyperref}
\usepackage[capitalize,nameinlink]{cleveref}
\usepackage{xcolor}
\usepackage{colortbl}
\usepackage{tcolorbox}
\usepackage[aboveskip=2pt]{subcaption}

\crefname{section}{Sec.}{Secs.}
\crefname{proposition}{Prop.}{Props.}
\crefname{lemma}{Lem.}{Lems.}
\crefname{model}{Mod.}{Mods.}
\crefname{appendix}{App.}{Apps.}

\usepackage{algorithm}
\usepackage{algorithmic}

\newlength\figureheight
\newlength\figurewidth

\usepackage{bm}
\newcommand{\mathbold}[1]{\bm{#1}}
\newcommand{\mbf}[1]{\mathbf{#1}}

\usepackage{xspace}
\newcommand{\eg}{\textit{e.g.}\xspace}

\newcommand{\etc}{\textit{etc.}\xspace}

\newcommand{\T}{\top}    %
\newcommand{\dd}{\,\mathrm{d}} %
\newcommand{\R}{\mathbb{R}}    %
\newcommand{\N}{\mathrm{N}}   %

\DeclareMathOperator{\E}{E}
\DeclareMathOperator{\Cov}{Cov}

\newcommand{\KL}[2]{\mathrm{D}_\mathrm{KL}\left[#1\,\|\,#2\right]}

\newcommand{\diff}{\,\mathrm{d}}

\newcommand{\vbeta}[0]{\mathbold{\beta}}

\newcommand{\vmu}[0]{\mathbold{\mu}}

\newcommand{\vxi}[0]{\mathbold{\xi}}

\newcommand{\vkappa}[0]{\mathbold{\kappa}}

\renewcommand{\mid}[0]{\,|\,}
\newcommand{\bigO}{\mathcal{O}}

\newcommand{\ve}{\mbf{e}}
\newcommand{\vf}{\mbf{f}}
\newcommand{\vg}{\mbf{g}}

\newcommand{\vm}{\mbf{m}}

\newcommand{\vp}{\mbf{p}}

\newcommand{\vv}{\mbf{v}}
\newcommand{\vw}{\mbf{w}}
\newcommand{\vx}{\mbf{x}}
\newcommand{\vy}{\mbf{y}}
\newcommand{\vz}{\mbf{z}}
\newcommand{\MA}{\mbf{A}}

\newcommand{\MF}{\mbf{F}}

\newcommand{\MI}{\mbf{I}}

\newcommand{\MK}{\mbf{K}}
\newcommand{\ML}{\mbf{L}}

\newcommand{\MP}{\mbf{P}}
\newcommand{\MQ}{\mbf{Q}}

\newcommand{\ODEtwoVAE}{ODE\textsuperscript{2}VAE\xspace}

\definecolor{cgray}{gray}{0.4}

\renewcommand{\paragraph}[1]{{\bf #1}~~}

\definecolor{mycolor0}{rgb}{0.2667,0.4471,0.7098}
\definecolor{mycolor1}{rgb}{0.1647,0.6706,0.3804}
\definecolor{mycolor2}{rgb}{0.8275,0.2627,0.3059}
\definecolor{mycolor3}{rgb}{0.5216,0.4392,0.7176}
\definecolor{mycolor4}{rgb}{0.8118,0.7255,0.4118}
\definecolor{mycolor5}{rgb}{0.2745,0.7176,0.8157}

\definecolor{mylcolor0}{rgb}{0.6902,0.7686,0.8863}
\definecolor{mylcolor1}{rgb}{0.5451,0.8902,0.6941}
\definecolor{mylcolor2}{rgb}{0.9412,0.7490,0.7647}
\definecolor{mylcolor3}{rgb}{0.8627,0.8392,0.9176}
\definecolor{mylcolor4}{rgb}{0.9569,0.9373,0.8667}
\definecolor{mylcolor5}{rgb}{0.7529,0.9020,0.9373}
\definecolor{mylcolor6}{rgb}{0.8750,0.8750,0.8750}
\definecolor{mygrey}{rgb}{0.93, 0.93, 0.93}

\definecolor{color2}{rgb}{0.75,0.75,0}
\definecolor{color1}{rgb}{1,0.498039215686275,0.0549019607843137}
\definecolor{color0}{rgb}{0.12156862745098,0.466666666666667,0.705882352941177}

\usepackage{tabularx}
\usepackage{array}
\newcommand{\PreserveBackslash}[1]{\let\temp=\\#1\let\\=\temp}
\newcolumntype{C}[1]{>{\PreserveBackslash\centering}p{#1}}

\newcommand{\nipstitle}[1]{{%
    \def\toptitlebar{\hrule height4pt \vskip .25in \vskip -\parskip} 
    \def\bottomtitlebar{\vskip .29in \vskip -\parskip \hrule height1pt \vskip .09in} 
    \phantomsection\hsize\textwidth\linewidth\hsize%
    \vskip 0.1in%
    \toptitlebar%
    \begin{minipage}{\textwidth}%
        \centering{\LARGE\bf #1\par}%
    \end{minipage}%
    \bottomtitlebar%
    \addcontentsline{toc}{section}{#1}%
}}

\usepackage{tikz,pgfplots,pgfmath}
\usetikzlibrary{plotmarks,shapes.arrows,calc,decorations.text}
\usetikzlibrary{decorations.pathreplacing}
\usetikzlibrary{patterns,shapes}
\pgfplotsset{compat=newest}

\pgfplotsset{every axis/.append style={
		grid style={line width=0.6pt,dotted,gray}}}

\pgfplotsset{every axis/.append style={
		legend style={inner xsep=1pt, inner ysep=0.5pt, nodes={inner sep=1pt, text depth=0.1em},draw=none,fill=none}
}}

\usepgfplotslibrary{groupplots}

\hypersetup{bookmarks=true}

\pgfplotsset{ignore legend/.style={every axis legend/.code={\let\addlegendentry\relax}}}

\usepackage{comment}

\usepackage{wrapfig}

\title{Scalable Inference in SDEs by Direct Matching of the Fokker--Planck--Kolmogorov Equation}

\author{%
  Arno Solin \\
  Aalto University\\
  Espoo, Finland\\
  \texttt{arno.solin@aalto.fi}
  \And
  Ella Tamir\\
  Aalto University\\
  Espoo, Finland\\
  \texttt{ella.tamir@aalto.fi}
  \And
  Prakhar Verma\\
  Aalto University\\
  Espoo, Finland\\
  \texttt{prakhar.verma@aalto.fi}
}

\begin{document}

\maketitle

\begin{abstract}
  Simulation-based techniques such as variants of stochastic Runge--Kutta are the {\it de facto} approach for inference with stochastic differential equations (SDEs) in machine learning. These methods are general-purpose and used with parametric and non-parametric models, and neural SDEs. Stochastic Runge--Kutta relies on the use of sampling schemes that can be inefficient in high dimensions. We address this issue by revisiting the classical SDE literature and derive direct approximations to the (typically intractable) Fokker--Planck--Kolmogorov equation by matching moments. We show how this workflow is fast, scales to high-dimensional latent spaces, and is applicable to scarce-data applications, where a non-parametric SDE with a driving Gaussian process velocity field specifies the model. 
\end{abstract}

\setlength{\columnsep}{5pt}
\setlength{\intextsep}{0pt}
\begin{wrapfigure}{r}{.4\textwidth}
  \vspace*{-2.5em}
  \hspace*{-4pt}%
  \resizebox{.4\textwidth}{!}{%
  \setlength{\figurewidth}{.56\textwidth}
  \setlength{\figureheight}{\figurewidth}
  \footnotesize
  \begin{tikzpicture}[outer sep=0]
      \newcommand{\splitfig}[2]{\tikz{%
      \draw[draw=white,thick,inner sep=0,path picture={
        \node [anchor=west,text width=\figurewidth] at (path picture bounding box.west){
          \includegraphics[width=\figurewidth]{#1}
        };}] (0,0) -- (.34\figurewidth,0) -- 
         (.54\figurewidth,\figureheight) -- (0,\figureheight) -- (0,0);
      \draw[draw=white,thick,inner sep=0,path picture={
        \node [anchor=east,text width=\figurewidth] at (path picture bounding box.east){
          \includegraphics[width=\figurewidth]{#2}
        };}] (.35\figurewidth,0) -- (\figurewidth,0) -- (\figurewidth,\figureheight) --
         (.55\figurewidth,\figureheight) -- (.35\figurewidth,0);}}

      \newcommand{\trifig}[3]{{%
      \draw[draw=white,thick,inner sep=0,path picture={
        \node [anchor=west,text width=\figurewidth] at (path picture bounding box.west){
          \includegraphics[width=\figurewidth]{#1}
        };}] (-0.0125\figurewidth,1.0000\figureheight) -- (0.1988\figurewidth,1.0000\figureheight) -- (0.4875\figurewidth,0.5000\figureheight) -- (0.1988\figurewidth,0.0000\figureheight) -- (-0.0125\figurewidth,0.0000\figureheight) -- (-0.0125\figurewidth,1.0000\figureheight);
      \draw[draw=white,thick,inner sep=0,path picture={
        \node [anchor=east,text width=\figurewidth] at (path picture bounding box.south east){
          \includegraphics[width=\figurewidth]{#2}
        };}] (1.0063\figurewidth,1.0108\figureheight) -- (1.0063\figurewidth,0.5108\figureheight) -- (0.5062\figurewidth,0.5108\figureheight) -- (0.2176\figurewidth,1.0108\figureheight) -- (1.0063\figurewidth,1.0108\figureheight);
      \draw[draw=white,thick,inner sep=0,path picture={
        \node [anchor=east,text width=\figurewidth] at (path picture bounding box.north east){
          \includegraphics[width=\figurewidth]{#3}
        };}] (1.0063\figurewidth,-0.0108\figureheight) -- (0.2176\figurewidth,-0.0108\figureheight) -- (0.5062\figurewidth,0.4892\figureheight) -- (1.0063\figurewidth,0.4892\figureheight) -- (1.0063\figurewidth,-0.0108\figureheight);
        }}  

  \trifig{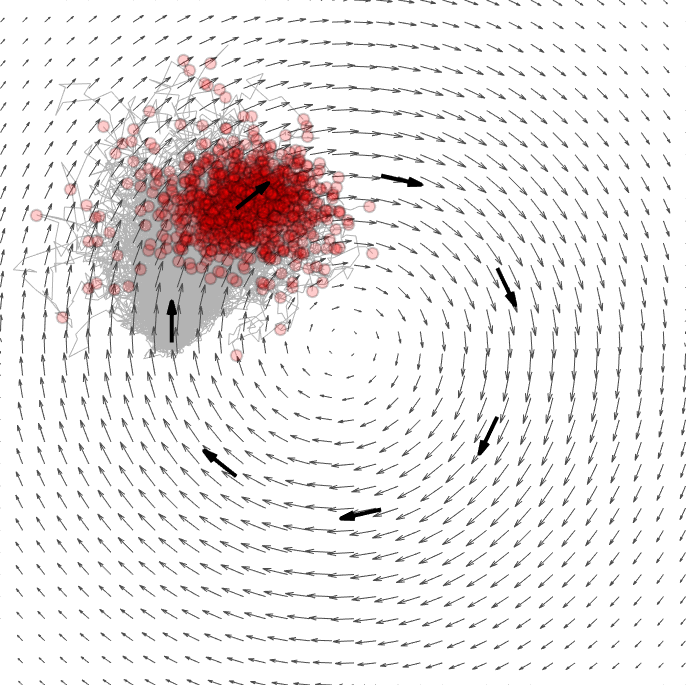}{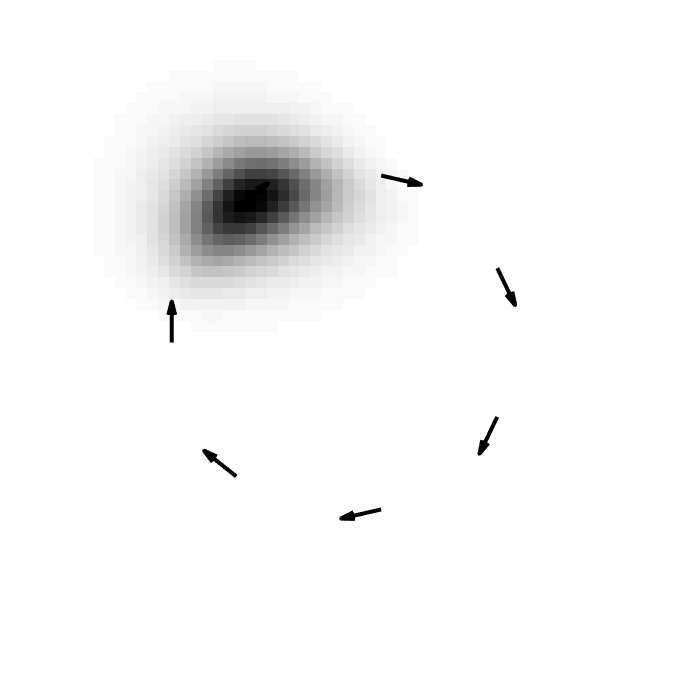}{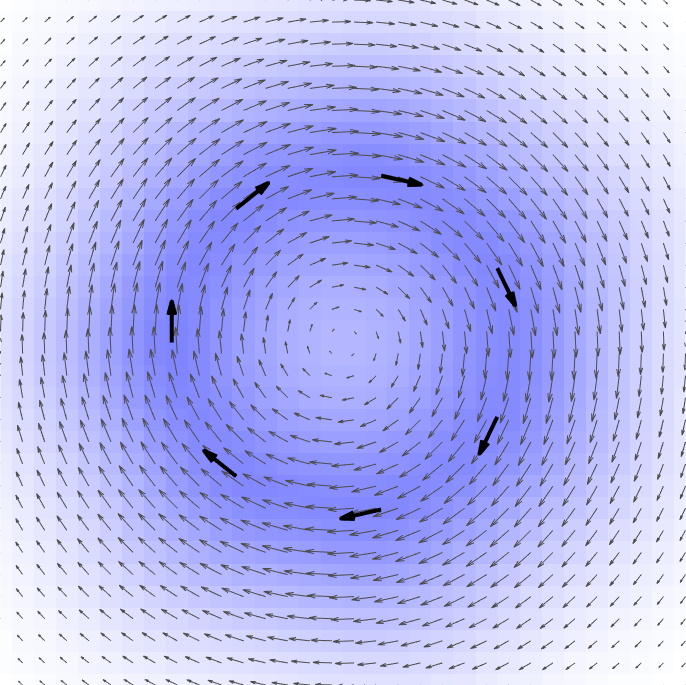}

  \tikzstyle{info} = [draw=none,fill=white,text width=2.7cm,inner sep=4pt,rounded corners=2pt,opacity=.8,align=center]
  \node[info] at (1.8,3.1) {1.\ Euler--Maruyama sample trajectories};
  \node[info] at (6.5,6.5) {3.\ Probability density $p(\vz,t)$ as solution to the Fokker--Planck--Kolmogorov PDE};  
  \node[info] at (2,7.2) {2.\ Gaussian approximation};    
  \node[info] at (6.2,1) {GP-SDE \\ (SDE model specification by a GP prior)};    

  \node[ellipse,minimum width=2.9cm,minimum height=2cm,rotate=25,draw=blue,thick] at (2.9,5.7) {}; 
  \node[circle,fill=blue,inner sep=3pt] at (2.9,5.7) {}; 
  
  \end{tikzpicture}}
  \caption{Views into solutions to SDEs.}%
  \label{fig:teaser}
\end{wrapfigure}
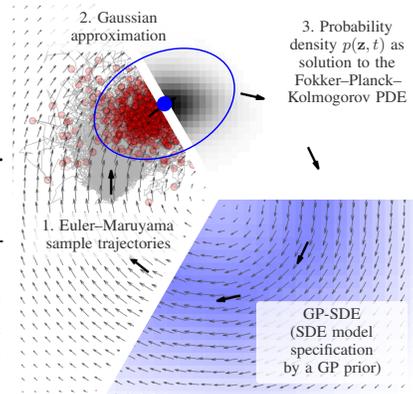

\section{Introduction}
Differential equations are the standard method of modelling {\em change} over time. In  deterministic systems the dynamics specifying how the system evolves, are typically written in the form of an ordinary differential equation (ODE). The dynamics act as prior knowledge and often stem from first-principles in application areas such as physics, control engineering, chemistry, or compartmental models in epidemiology and pharmacokinetics. Recently, learning ODE dynamics with modern automatic differentiation packages in machine learning has awakened an interest in black-box learning of continuous-time dynamics (\eg, \cite{chen2018neuralode,rubanovalatentode2019}) and enabled their more general use across time-series modelling applications.

A stochastic differential equation (SDE, \cite{Oksendal:2003,Sarkka+Solin:2019}) can be seen as a generalization of ODEs to stochastic dynamical settings, where the driving forces fluctuate or are uncertain. Stochastic dynamics appear naturally in applications where small (and typically unobserved) forces interact with the process, such as tracking applications, molecule motion, gene modelling, or stock markets. In machine learning, SDE models have received wide-spread attention due to their robustness and appealing properties for uncertainty quantification.

The concept of a `solution' to an SDE is broader than that of an ODE. As the process is stochastic, the full solution entails a probability distribution, $p(\vz,t)$, depending on time $t$ and covering the space $\vz$ (see, \eg, \cite{Rogers+Williams:2000b}). For It\^o type SDEs, the evolution of the probability mass can be described in terms of the Fokker--Planck--Kolmogorov (FPK) partial differential equation (backward Kolmogorov equation). This equation is typically intractable, and instead the {\it de~facto} approach for inference in SDEs in machine learning is sampling. The most common approaches in this space are based on Stochastic Runge--Kutta schemes (such as the {\em Euler--Maruyama scheme}) which are derived from the It\^o--Taylor series. These schemes sample realization trajectories of the SDE by driving the dynamics with numerical simulation of Brownian motion.
However, these schemes suffer from drawbacks both related to (ordinary) Runge--Kutta methods---such as step-size and sensitivity to stiffness---as well as problems associated with any sampling schemes, such as a high number of samples required for an accurate representation of the underlying distribution.

Despite these problems, few contemporary SDE approaches in machine learning explore SDE solutions beyond stochastic Runge--Kutta (or even the Euler--Maruyama scheme). Our aim is to try to broaden this view, and in \cref{fig:teaser} we sketch an example where we show three alternative solution perspectives to a Gaussian process prior SDE model (GP-SDE): the FPK probability density field, Euler--Maruyama samples, and a Gaussian assumed density approximation. We argue that a Gaussian approximation in latent space SDEs is reasonable, as Gaussian approximations are typically employed anyway in observation models, and allow for speeding-up learning by an order of magnitude.

The contributions of this paper are as follows.
{\em (i)}~We go through the workflow connecting `random ODE' models with It\^o SDEs driven by a Gaussian process prior over the velocity field, which allows for convenient specification of prior knowledge on the vector field and induces an implicit prior over the SDE trajectories;
{\em (ii)}~We revisit the classical SDE literature and derive direct approximations to the (typically intractable) Fokker--Planck--Kolmogorov equation in an assumed density Gaussian form that avoids sampling-based inference in the latent space, which makes inference fast and does not require sampling a high number of trajectories;
{\em (iii)}~We show how this workflow is fast, applicable to scarce-data applications, and how it also extends to previously presented latent SDE models.

\subsection{Related Work}
Neural ODEs \cite{chen2018neuralode} model ODE dynamics by a neural network. 
Such models were developed further in \cite{rubanovalatentode2019}, where the encoder is an ODE-RNN that improves modelling of irregularly sampled time series.
A latent Bayesian neural ODE model, \ODEtwoVAE, was examined in \citet{yildiz2019ode2vae}, where an encoder is combined with an ODE model whose second order dynamics are given by a Bayesian neural network. %
The neural ODE paradigm of modelling latent dynamics has been expanded to neural SDEs \cite{li2020sdeadjoint, jung-su2018adaptive,tzen2019neuralsde, kidger2021sdegan}, where the typical workflow is that a variational autoencoder (VAE, \cite{kingma2014autoencoding,rezende2014stochastic}) is combined with a latent neural SDE, whose drift and diffusion are modelled by neural networks. 
In addition to modelling time series, neural SDEs have been used in generative models~\cite{ho2020denoise,song2019genlang,song2020sdegen}, where the generation of images from noise is modelled as the reverse-time process of a diffusion SDE by using Langevin dynamics on score-based models. Continuous normalizing flows are another model family, which applies ODE dynamics in a generative model \cite{chen2018neuralode, grathwohl2018scalable}. 

These works leverage simulation/sampling for solving the SDE in the latent space. The model can be trained by a stochastic adjoint method \cite{li2020sdeadjoint,kidger2020neural}. More recently, latent neural SDEs have been trained deterministically by moment matching \cite{look2020deterministic}. However, they discretized the system before matching the moments, while we form a direct approximation to the solution of the FPK. Compared to optimizing the moment ODEs, as discussed in \cite{look2020deterministic}, by maximizing likelihood, we regularize during inference by Gaussian process priors, or prior stochastic processes as in \cite{li2020sdeadjoint}.
Approximative solutions to non-linear SDEs have been applied earlier in filtering theory, where the optimal filter is approximated by a Gaussian assumed density filter \citep{Kushner:1967}. In \cite{Sarkka+Sarmavuori:2012}, such approximations are used for  continuous-discrete state-space modelling. An alternative to assumed density filters are local linearization methods \cite{ozaki1992locallin,shoji1998locallin}, and simulation-based It\^o--Taylor series solutions, stochastic Runge--Kutta methods, and leapfrog methods such as Verlet for second-order SDEs (see \cite{kloeden2012numerical, Sarkka+Solin:2019}).
The approximations presented in this work are also related to GP approximations \cite{Archambeau+Opper:2011,Ala-Luhtala:2015} of SDEs. The linearization approximations are related to statistical linearization 
\cite{Gelb:1974,Socha:2008}, and variational approximations \cite{Archambeau+Opper:2011}.

Orthogonally to the SDE inference, we also consider SDE model specification in terms of GP priors. The seminal work by \citet{Ruttor+Batz+Opper:2013} considered GP-SDE models with unit diffusion. \citet{yildiz2018sdegp} built a model, where the drift and diffusion are sparse Gaussian processes with time-independent kernels. In \citet{hedge2019deepgp}, a spatio-temporal SDE with GP priors for the drift and diffusion was combined with a GP as a continuous version of deep Gaussian processes. 
State-space models with a GP latent state transition function \cite{frigola2014gpss} train a non-parametric latent process to approximate unobserved dynamics. These are related to hierarchical GP dynamics \cite{titsias10latentgp, lawrence2007gphierarchy}, where prior knowledge of the system can be encoded in multi-level hierarchies, for  modelling, \eg, walking dynamics.

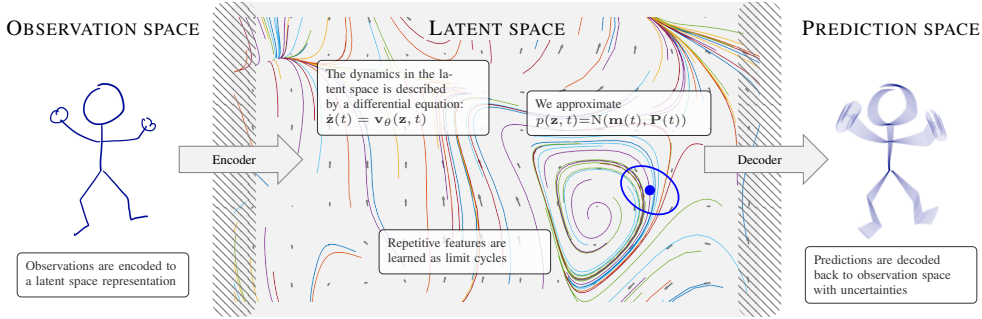
\begin{figure*}[!t]
  \centering\tiny
  \setlength{\figurewidth}{1.22\textwidth}
  \resizebox{\textwidth}{!}{%
  \begin{tikzpicture}[outer sep=0]

    \node[minimum width=.25\figurewidth,minimum height=.3\figurewidth,fill=black!0,
      rounded corners=6pt] at (.125\figurewidth,0) {\includegraphics[width=.1\figurewidth]{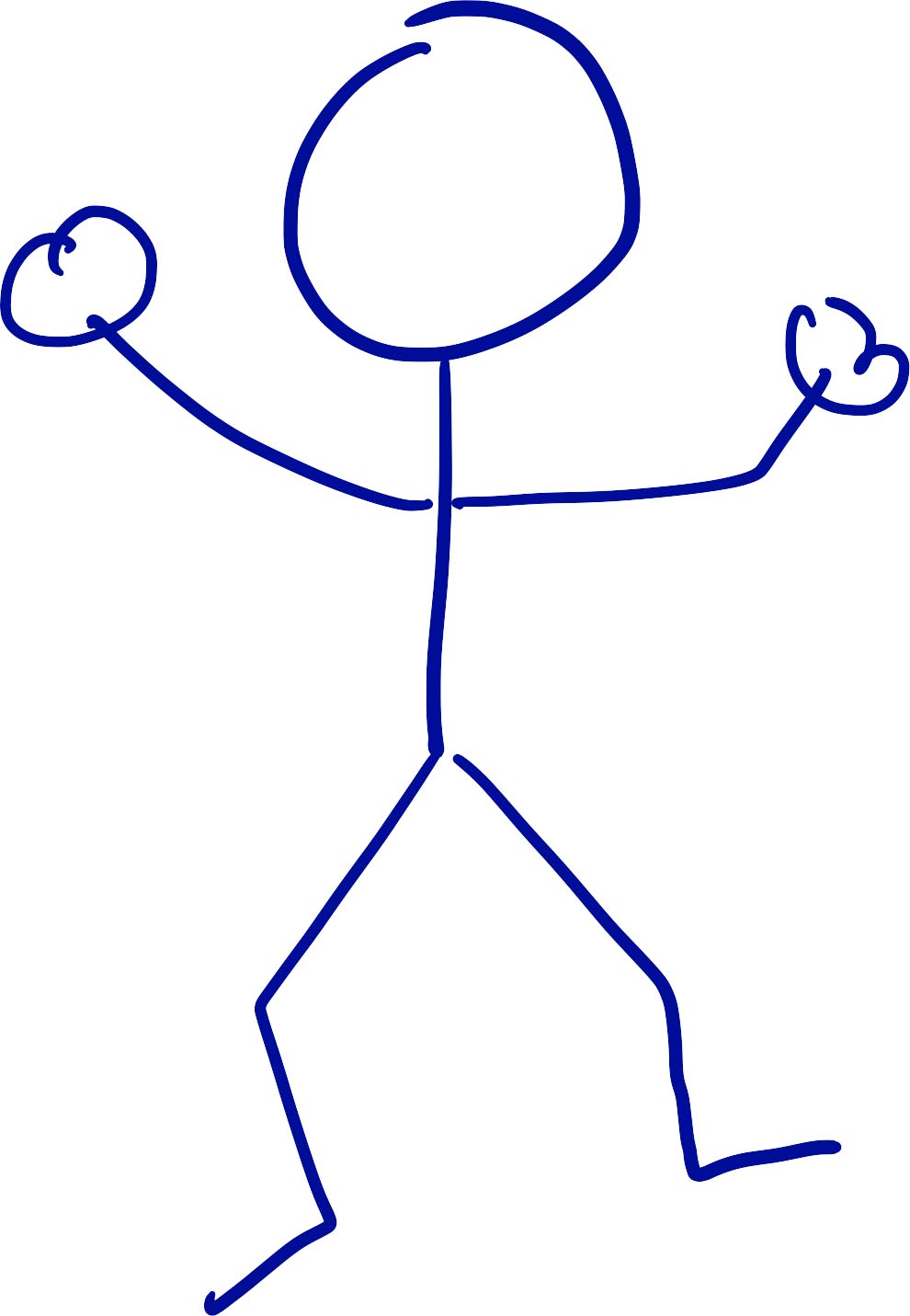}};
    
    \node[minimum width=.5\figurewidth,minimum height=.3\figurewidth,fill=black!5,
      rounded corners=6pt] at (.5\figurewidth,0) {\includegraphics[width=.5\figurewidth]{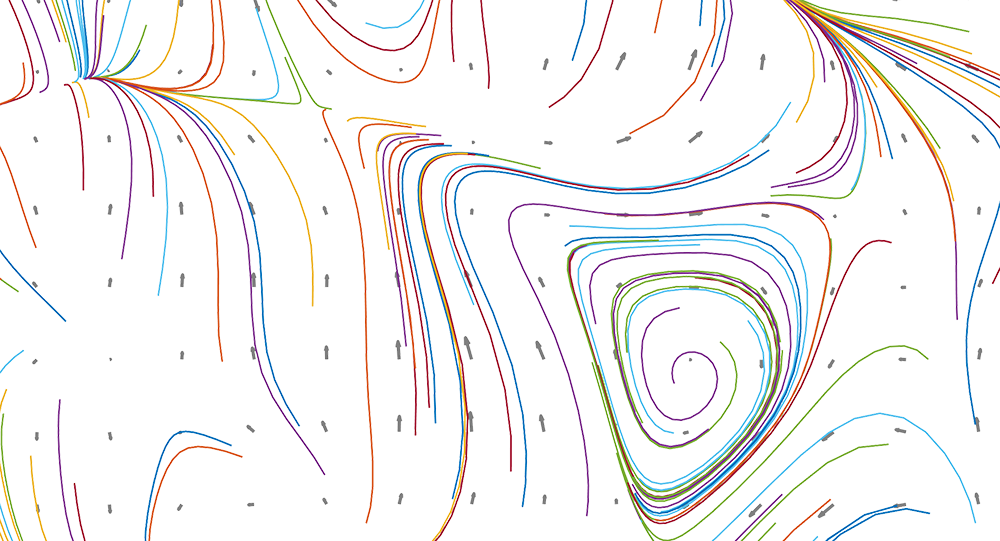}};     
    
    \node[minimum width=.25\figurewidth,minimum height=.3\figurewidth,fill=black!0,
      rounded corners=6pt] at (.875\figurewidth,0) {\includegraphics[width=.12\figurewidth]{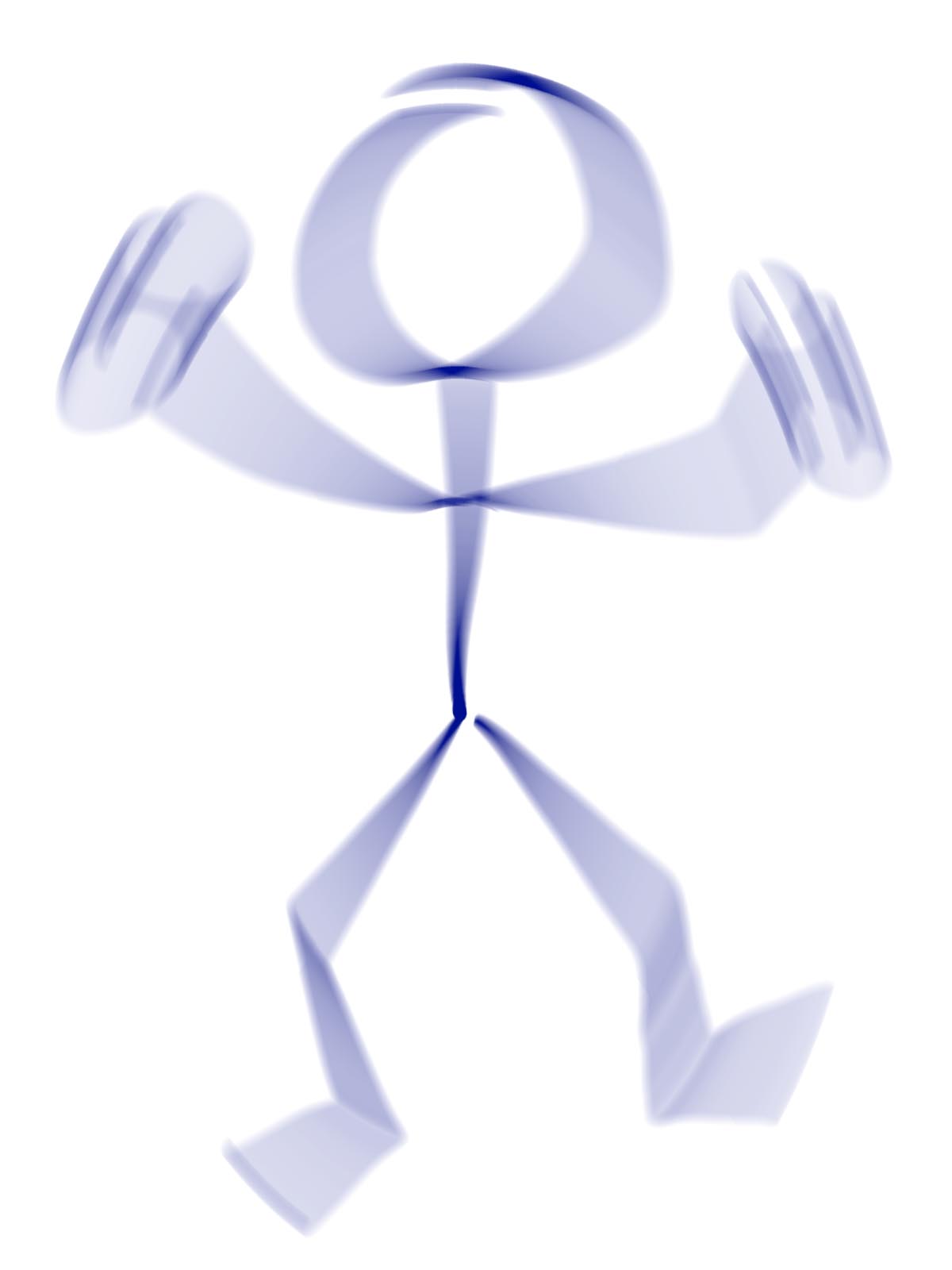}};

    \node[minimum width=7mm,minimum height=.3\figurewidth,
      rounded corners=6pt, pattern=north west lines, pattern color=black!50] 
      at (.75\figurewidth,0) {};
    \node[minimum width=7mm,minimum height=.3\figurewidth,
      rounded corners=6pt, pattern=north west lines, pattern color=black!50] 
      at (.25\figurewidth,0) {};    

    \tikzstyle{arrow} = [draw=black!10, single arrow, minimum height=20mm, minimum width=8mm, single arrow head extend=1mm, fill=black!5, draw=black!50, anchor=center, rotate=0, inner sep=1pt]
    \node[arrow] at (.25\figurewidth,0) {Encoder};
    \node[arrow] at (.75\figurewidth,0) {Decoder};      

    \node at (.125\figurewidth,.125\figurewidth) {\normalsize\sc Observation space};
    \node[fill=black!5] at (.500\figurewidth,.125\figurewidth) {\normalsize\sc Latent space};    
    \node at (.875\figurewidth,.125\figurewidth) {\normalsize\sc Prediction space};    

    \tikzstyle{info} = [draw=black,fill=white,text width=2.5cm,inner sep=4pt,rounded corners=2pt,opacity=.8]
    \node[info] at (.125\figurewidth,-.11\figurewidth) {Observations are encoded to a latent space representation};    
    \node[info] at (7,1) {The dynamics in the latent space is described by a differential equation: $\dot{\vz}(t) = \vv_\theta(\vz,t)$};
    \node[info] at (8,-1.5) {Repetitive features are learned as limit cycles};
    \node[info,text width=2.7cm] at (10.5,.75) {We approximate $p(\vz,t) {=} \mathrm{N}(\vm(t),\MP(t))$};        
    \node[info] at (.875\figurewidth,-.11\figurewidth) {Predictions are decoded back to observation space with uncertainties};            

     \node[circle,draw=blue,fill=blue,inner sep=1.5pt] at (11,-.5) {};     
     \node[ellipse,draw=blue,thick,minimum width=1cm,minimum height=.7cm,rotate=-30] at (11,-.5) {};
        
  \end{tikzpicture}}
  \caption{Latent dynamics workflow. The observations (left) are encoded into the latent space, where the dynamics of the system are learned as an SDE model. We approximate the solution to the SDE by a Gaussian for which we can approximate the dynamics of the first moments by an ODE system, thus avoiding sampling in the latent space. Predictions are finally mapped back.}
  \label{fig:workflow}
\end{figure*}

\section{Methods}
\label{sec:methods}
We draft the methodology based on the latent dynamics workflow components as presented in \cref{fig:workflow}. The focus is first on specifying models for the latent space dynamics, starting from implicit priors in terms of random ODEs which we frame as It\^o SDEs. Thereafter, the focus shifts from model specification to inference, where we show that the Fokker--Planck--Kolmogorov equation can be efficiently approximated in an assumed density form, and finally brings us to cover the likelihood structure of these models as well. 

\subsection{Random Field Ordinary Differential Equations as SDEs}
\label{sec:gp-sde}
The continuous dynamics of a latent (unobserved) $\vz(t)\in\R^d$ can conveniently be written in the form of a general first-order ordinary differential equation
\begin{equation}\label{eq:ode}
  \frac{\diff}{\diff t} \vz(t) = \vv_\theta(\vz(t),t),
\end{equation}
where $\vv_\theta(\cdot): \R^d \times \R_+ \to \R^d$ denotes the velocity field parametrized by $\theta$. This is a general form of a non-linear ODE system, where the dynamics are deterministic and fully characterized by $\vv_\theta(\cdot)$. The methodology presented in this section directly extends to the case where the vector field $\vv_{\theta}$ is time-dependent, but we omit the time dimension for simplicity of notation. 
Previously, the implicit prior on $\vz(t)$ over $t$ specified by \cref{eq:ode} has been generalized to stochastic models by considering $\vv_\theta(\cdot)$ to be stochastic. These models are known as `random' ODE models, and the random field $\vv_\theta(\cdot)$ is typically either characterized by a Gaussian random field or Gaussian process model (see, \eg, \cite{Ruttor+Batz+Opper:2013,hedge2019deepgp}) or some parametric model (\eg, \cite{li2020sdeadjoint}). %

\begin{figure*}[t]
	\tiny
	\pgfplotsset{axis on top,scale only axis,width=\figurewidth,height=\figureheight, y  tick label style={rotate=90},xlabel={$z_1$},ylabel={$z_2$}}
	\setlength{\figurewidth}{.25\textwidth}
	\setlength{\figureheight}{\figurewidth}
	\centering
	\begin{subfigure}[t]{.3\textwidth}
		\centering
\begin{tikzpicture}

\begin{axis}[
height=\figureheight,
tick align=outside,
tick pos=left,
width=\figurewidth,
x grid style={white!69.01960784313725!black},
xmin=-1, xmax=1,
xtick style={color=black},
y grid style={white!69.01960784313725!black},
ymin=-1, ymax=1,
ytick style={color=black}
]
\addplot graphics [includegraphics cmd=\pgfimage,xmin=-1.33116883116883, xmax=1.26623376623377, ymin=-1.28571428571429, ymax=1.31168831168831] {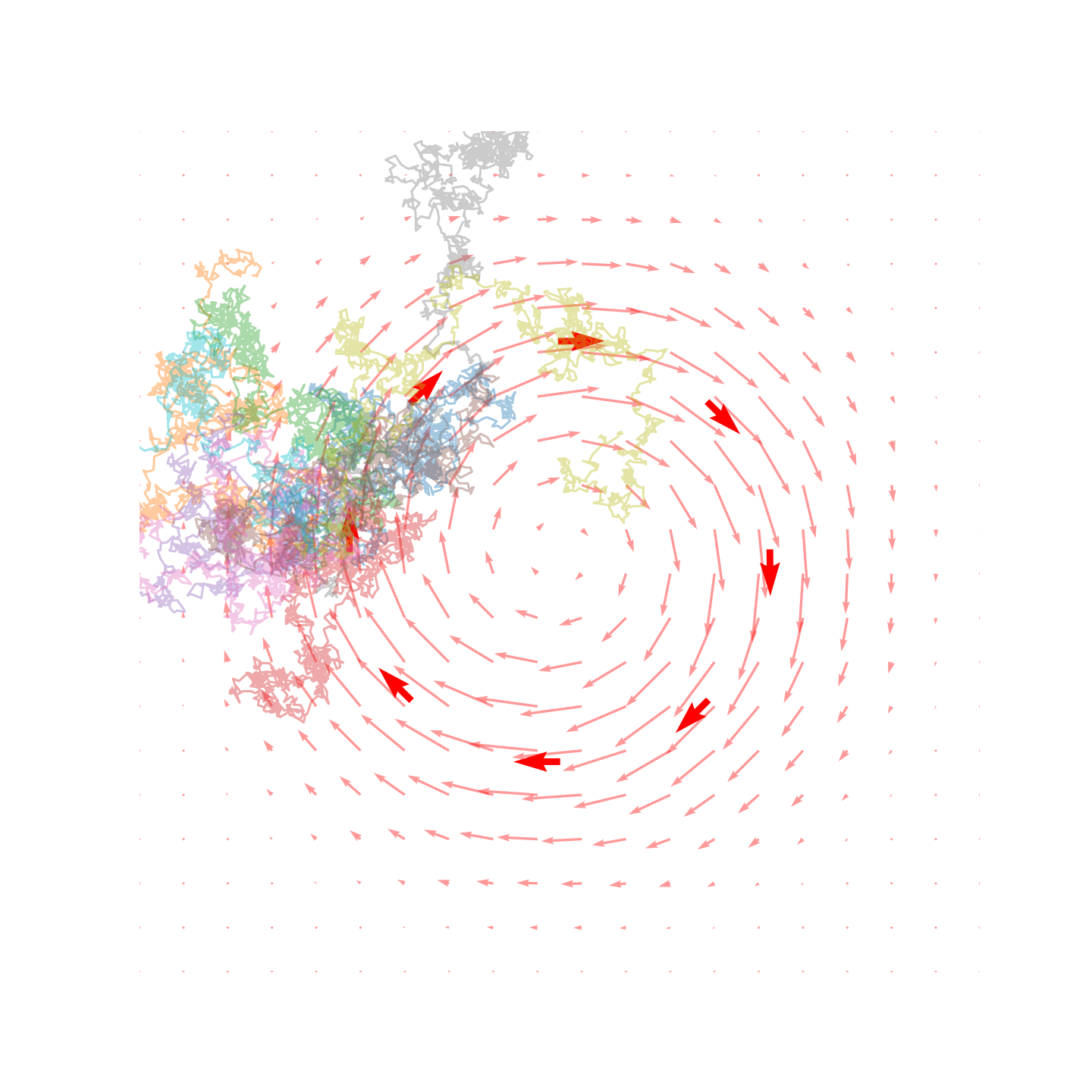};
\end{axis}

\end{tikzpicture}\\[0em]
		\caption{Independent RBF priors}
	\end{subfigure}
    \hfill
	\begin{subfigure}[t]{.3\textwidth}
		\centering
\begin{tikzpicture}

\begin{axis}[
height=\figureheight,
tick align=outside,
tick pos=left,
width=\figurewidth,
x grid style={white!69.01960784313725!black},
xmin=-1, xmax=1,
xtick style={color=black},
y grid style={white!69.01960784313725!black},
ymin=-1, ymax=1,
ytick style={color=black}
]
\addplot graphics [includegraphics cmd=\pgfimage,xmin=-1.33116883116883, xmax=1.26623376623377, ymin=-1.28571428571429, ymax=1.31168831168831] {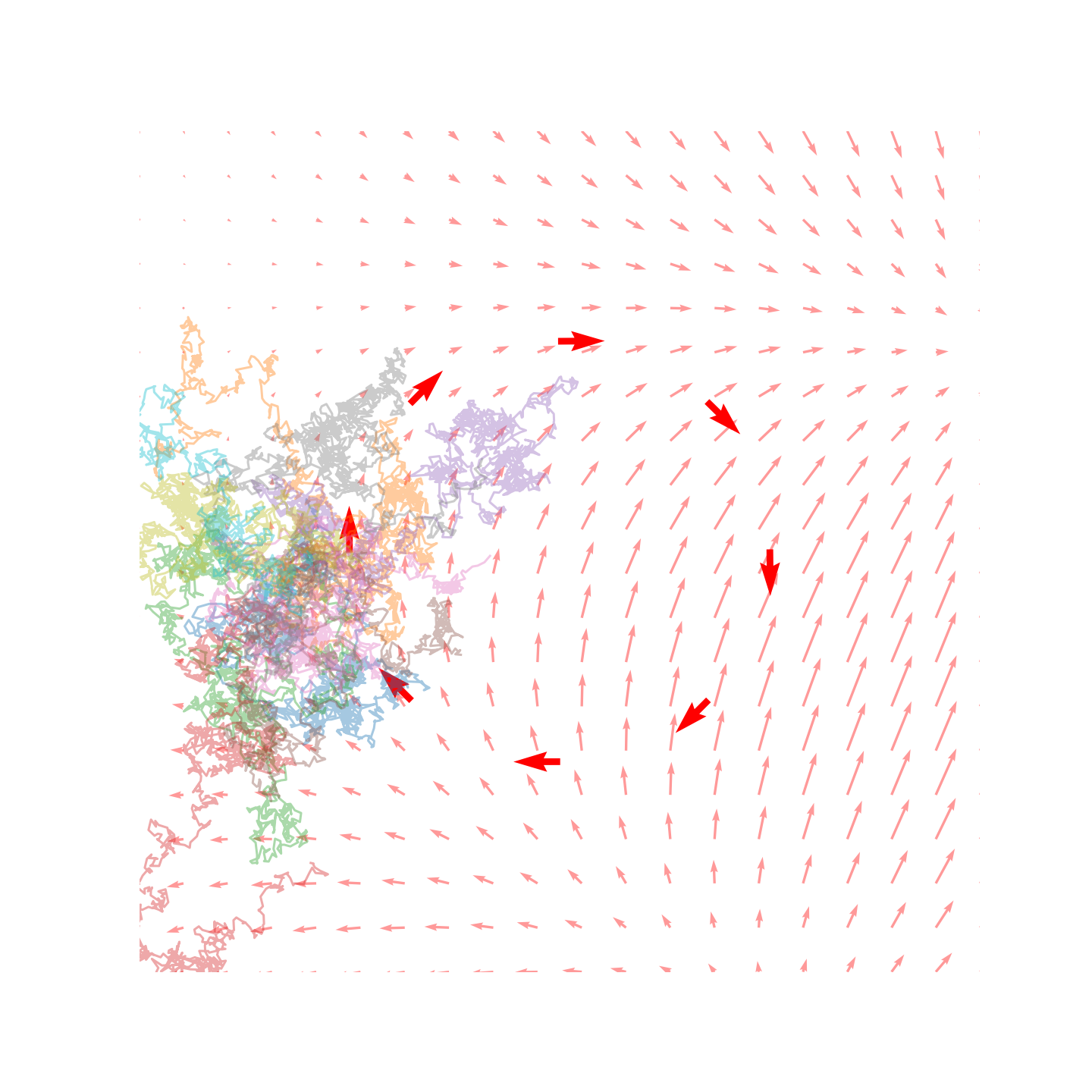};
\end{axis}

\end{tikzpicture}\\[0em]
	    \caption{Curl-free prior}
	\end{subfigure}
	\hfill
	\begin{subfigure}[t]{.3\textwidth}
		\centering
\begin{tikzpicture}

\begin{axis}[
height=\figureheight,
tick align=outside,
tick pos=left,
width=\figurewidth,
x grid style={white!69.01960784313725!black},
xmin=-1, xmax=1,
xtick style={color=black},
y grid style={white!69.01960784313725!black},
ymin=-1, ymax=1,
ytick style={color=black}
]
\addplot graphics [includegraphics cmd=\pgfimage,xmin=-1.33116883116883, xmax=1.26623376623377, ymin=-1.28571428571429, ymax=1.31168831168831] {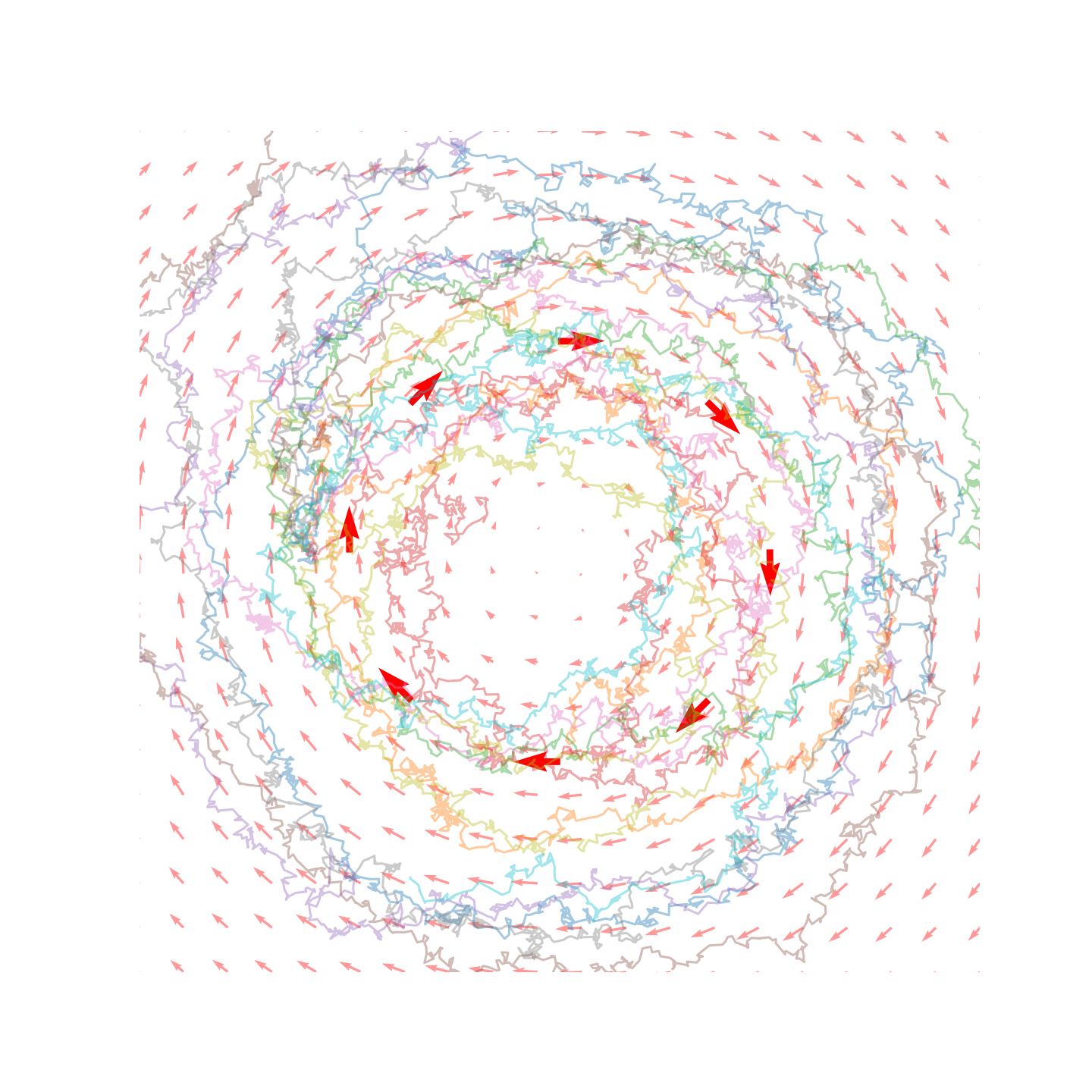};
\end{axis}

\end{tikzpicture}\\[0em]
	    \caption{Divergence-free prior}  
	\end{subfigure}
	\caption{Effect of different GP priors in an GP-SDE with 8 observations (large red arrows) to the GP posterior (small red arrows): (a)~Shows results for independent RBF priors over $z_1$ and $z_2$; (b)~shows results for the curl-free covariance function (encoding {`loop-aversion'}); (c)~shows results for the divergence-free covariance function (encoding {`energy preservation'}). The hyperparameters ($\ell=0.2, \sigma^2=0.1$) are the same in each.}
	\label{fig:effect-of-prior}
\end{figure*}

We consider an unconventional ODE model (or actually no ODE model at all, to be precise), where we specify a GP prior \citep{Rasmussen+Williams:2006} over the velocity field in form of a multi-output Gaussian process prior:
\begin{equation}\label{eq:gpode}
  \vv(\vz,t) \sim \mathrm{GP}(\vmu(\vz),\vkappa(\vz,\vz')),
\end{equation}
where $\vmu: \R^d \to \R^d$ is a mean function and $\vkappa: \R^d {\times} \R^d \to \R^{d{\times}d}$ is a matrix-valued covariance function. The Gaussian process prior is completely specified by its mean and covariance function, which encapsulate the assumptions about the sample processes/fields $\vv$ (such as continuity, differentiability, curl, divergence, \etc):
  $\vmu(\vz) := \mathrm{E}[\vv(\vz)]$ and
  $\vkappa(\vz,\vz') := \mathrm{E}[(\vv(\vz)-\vmu(\vz))(\vv(\vz')-\vmu(\vz'))^*]$.
In \cref{fig:effect-of-prior}, we will consider examples of useful vector-valued covariance functions that encode properties on the vector field.
For inference, the GP is conditioned on input--output pair observations $\mathcal{D} = \{(\vz_i,\Delta\vz_i)\}_{i=1}^n$ of the vector field, where $\Delta\vz_i$ represents the observed derivative at $\vz_i$. The conditioned vector field representation for an arbitrary point $\vz_*$ in the latent space can be given by	%
\begin{equation}
  \vv(\vz_*) \mid \mathcal{D} \sim \mathrm{GP}(\E[\vv(\vz)\mid \mathcal{D}], \Cov[\vv(\vz)\mid \mathcal{D}]),
\end{equation}
where the $\E[\cdot]$ and $\Cov[\cdot]$ denote the marginal mean and (co)variance (for the multi-output GP, which means that the marginals are vector-valued). These take the form \cite{Rasmussen+Williams:2006}:    
  $\E[\vv(\vz_*)] = \MK_* \hat\MK^{-1} \vy$ and
  $\Cov[\vv(\vz_{*})] = \vkappa(\vz_*,\vz_*) - \MK_* \hat\MK^{-1} \MK_*^\T$,
where $\vy$ are the stacked observations of the derivatives such that $\vy = (\Delta\vz_1^\T, \ldots, \Delta\vz_n^\T)^\T$. The Gram matrix $\MK$ corresponds to evaluations of the covariance function such that $\MK_{ij} = \vkappa(\vz_i,\vz_j)$ are sub-blocks of $\MK$ corresponding to the observation pairs $(i,j)$, and $\hat\MK = \MK + \gamma\MI$, where $\gamma$ is a nugget (observation noise/discrepancy) term, and $\MK_*$ is the cross-covariance between $z$ and $z_*$. The prohibitive cubic computational scaling associated with GP models manifests in the inversion of $\hat\MK$, and thus for large $n$, approximations based on inducing points or projections are used in practice to avoid this explicit inversion.
In the light of \cref{eq:ode} with $\vv(\vz) \sim \mathrm{GP}(\cdot, \cdot)$, the ODE is driven by a multi-dimensional Gaussian random field conditioned on $\mathcal{D}$. A straightforward way of dealing with a model of this kind, is to do inference by sampling random draws of the velocity field from the GP, and then drive the ODE with those samples (can be viewed as an Monte Carlo approach for drawing ODE realizations).

However, a more convenient way is to specify the prior over the stochastic dynamics in a stochastic differential equation form. At its core, a lot of previous work in this space hinges on the realization that if everything is essentially Gaussian, an equivalent model can be specified in terms of an It\^o SDE describing the stochastic evolution of trajectories affected by the GP velocity field \citep[see][for discussion]{hedge2019deepgp}. Informally, this takes the white noise form
  $\frac{\diff}{\diff t} \vz(t) = \vf(\vz) + \ML(\vz) \, \vw(t),$
where $\vf(\vz) = \E[\vv(\vz)]$ and $\ML(\vz)$ denotes a square-root factor such that $\ML \ML^\T = \Cov[\vv(\vz)]$ (in the scalar case, just the square-root, and in the multi-output case, \eg, the Cholesky factor). Here $\vw(t)$ is a white noise process with unit spectral density.
It is worth noting that we do not give guarantees for a direct link between the random ODE in \cref{eq:gpode} and the following SDE formulation (see \cref{app:on-the-construction} for discussion). Yet, formally we write a similarly-behaving SDE in the standard It\^o SDE  form:
\begin{equation}\label{eq:SDE}
  \diff \vz(t) = \vf(\vz,t) \diff t + \ML(\vz,t) \diff \vbeta(t),
\end{equation}
where $\diff \vbeta(t)$ is vector-valued unit Brownian motion (the spectral density $\MQ$ is set to $\MI$). %
For a GP-SDE, the drift is driven by the GP mean, $\vf(\vz,t) := \E[\vv(\vz)]$ and the diffusion by the square-root factor of the marginal covariance at $\vz(t)$, $\ML(\vz,t) := \sqrt{\Cov[\vv(\vz)]}$.  %
To be precise, the GP-SDE drift and diffusion at a point $\vz_{*}$ are determined by the GP predicted mean and variance at $\vz_{*}$, which can be written as
\begin{multline}\label{eq:drift-diffusion}
\vf(\vz_{*},t) {=} \E[\vv(\vz_{*})] {=} \MK_* \hat\MK^{-1} \vy \text{~~and~~}
\ML(\vz_{*},t) {=}  \sqrt{\Cov[\vv(\vz_{*})]} {=} \sqrt{\vkappa(\vz_*,\vz_*) - \MK_* \hat\MK^{-1} \MK_*^\T}.
\end{multline}

\subsection{Fokker--Planck--Kolmogorov Equation}
\label{sec:FPK}
\begin{figure*}[t]
  \centering\tiny
  \setlength{\figurewidth}{.12\textwidth}%
  \setlength{\figureheight}{\figurewidth}
  \begin{tikzpicture}[inner sep=0]

  \tikzstyle{arrow} = [draw=black!10, single arrow, minimum height=10mm, minimum width=3mm, single arrow head extend=1mm, fill=black!10, anchor=center, rotate=-90, inner sep=2pt]
  \tikzstyle{box} = []%

  \node[box] at ({-.2*\figurewidth},{0*\figureheight}) {\includegraphics[width=.95\figurewidth]{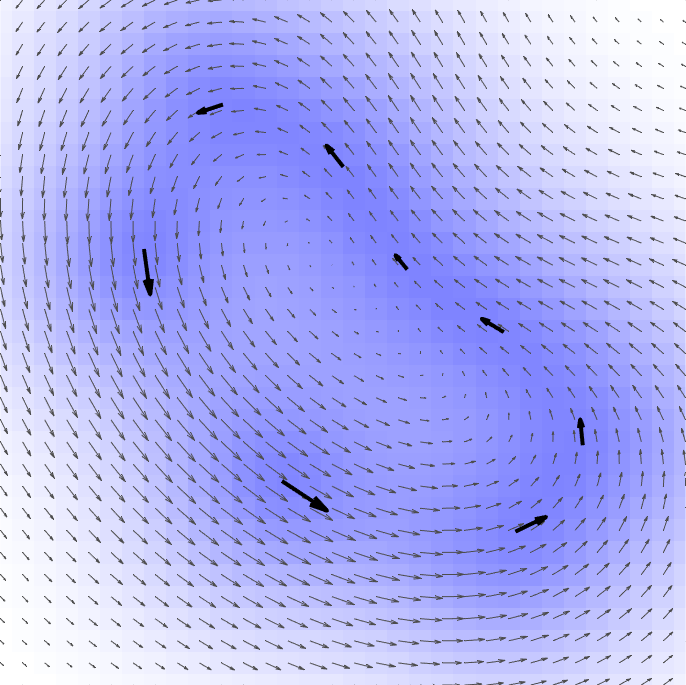}};

  \foreach \x [count=\i] in {10,50,100,200,300,400,500} {

     \node[box] at ({\figurewidth*\i},{0*\figureheight}) {\includegraphics[width=.95\figurewidth]{fig/bean-fpk-\x}};

     \node[box] at ({\figurewidth*\i},{-1*\figureheight}) {\includegraphics[width=.95\figurewidth]{fig/bean-em-\x.jpg}};     

     \node[box] at ({\figurewidth*\i},{-2*\figureheight}) {\includegraphics[width=.95\figurewidth]{fig/bean-ems-\x.jpg}};   

     \node[box] at ({\figurewidth*\i},{-3*\figureheight}) {\includegraphics[width=.95\figurewidth]{./fig/bean-adf-\x}};

  }

  \foreach \x [count=\i] in {1,2,3,4,5,6} {
    \draw[draw=black!50] ({\figurewidth*\i+\figurewidth/2},{0.5*\figureheight}) --
                         ({\figurewidth*\i+\figurewidth/2},{-3.5*\figureheight});
  }

  \foreach \x [count=\i] in {1,2,3} {
    \draw[draw=black!50] 
      ({0.5*\figureheight},{-\i*\figureheight+\figureheight/2}) --
      ({7.5*\figureheight},{-\i*\figureheight+\figureheight/2});
  }  

  \foreach \x [count=\i] in {0.1,0.5,1.0,2.0,3.0,4.0,5.0} {  
     \node[] at ({\figurewidth*\i},.55\figureheight) {$t = \x$~s};}
  \node[] at ({-.2\figurewidth},.55\figureheight) {GP velocity field};

  \node[rotate=90] at ({.45\figurewidth}, 0\figureheight) {Full FPK solution};
  \node[rotate=90] at ({.45\figurewidth},-1\figureheight) {Paths};
  \node[rotate=90] at ({.45\figurewidth},-2\figureheight) {Samples};
  \node[rotate=90] at ({.45\figurewidth},-3\figureheight) {Assumed density};

  \node[draw=black!90,inner sep=1pt] at ({-.2*\figurewidth},{-0.6\figureheight}) {\includegraphics[width=.8\figurewidth,height=2mm]{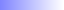}};
  \node[align=center] at ({-.2*\figurewidth},{-0.8\figureheight}) {\tiny GP marginal std};

  \node[box,inner sep=3pt] at ({-.2*\figurewidth},{-3\figureheight}) {
    \begin{minipage}{2cm}\tiny
      \tikz[baseline=-0.5ex]\draw[black,dashed] (0,0) -- (.3,0);~Exact Gaussian \\
      \tikz[baseline=-0.5ex]\draw[blue] (0,0) -- (.3,0);~Linearized \\
      \tikz[baseline=-0.5ex]\draw[red] (0,0) -- (.3,0);~Matched
    \end{minipage}};
            
  \end{tikzpicture}

  \caption{Approximations to the FPK equation: The {\bf top-left} figure shows the 8 observations (black arrows) and the inferred GP velocity (grey arrows, marginal uncertainty in shaded blue). The {\bf top-row} shows the progression of the probability mass $p(\vz,t)$ following the Fokker--Planck--Kolmogorov equation. The {\bf middle rows} show Euler--Maruyama sample trajectories for the problem, and the {\bf bottom row} compares the two assumed density approximations to the exact Gaussian approximation of the FPK solution. The bottom row ellipses are $95\%$ confidence regions. %
  }
  \label{fig:FPK-approx}
\end{figure*}

We are interested in {\em solving} SDE models of the form in \cref{eq:SDE}, but without the restriction that the drift and diffusion are defined by a Gaussian process, and present the related theory with a model-agnostic view on the problem. Because the resulting solutions are stochastic processes, the full solution to the SDE can be characterized by its time-evolving probability density function. Let $\vz(t_0) \sim p(\vz(t_0))$ be some initial condition which we assumed to be independent of the Brownian motion. The probability density $p(\vz,t)$ of the solution of the SDE in \cref{eq:SDE} solves the Fokker--Planck--Kolmogorov (FPK) partial differential equation (PDE):
\begin{multline}\label{eq:fpk}
  \frac{\partial p(\vz,t)}{\partial t} = 
  {-} \sum_i \frac{\partial}{\partial z_i}
  [f_i(\vz,t) \, p(\vz,t)] 
  {+} \frac{1}{2} \sum_{i,j} 
  \frac{\partial^2}{\partial z_i \, \partial z_j}
  \left\{ [\ML(\vz,t) \, \MQ \, \ML^\T(\vz,t)]_{ij} \, p(\vz,t) \right\}.
\end{multline}
For a proof, see \cite{Sarkka+Solin:2019}. This PDE is also known as the Fokker--Planck equation (in physics) and the forward Kolmogorov equation (in stochastics). An  appealing alternative form \cite[Sec.~5.3]{Sarkka+Solin:2019} of the FPK equation can be given in terms of the following evolution equation with the adjoint operator $\mathcal{A}^*$:
\begin{equation} \label{eq:fpk-op2}
  \frac{\partial p}{\partial t} = \mathcal{A}^* \, p, \text{~with~}
  \mathcal{A}^*(\bullet) =
  - \sum_i \frac{\partial}{\partial z_i}
  [f_i(\vz,t) \, (\bullet)] 
  + \frac{1}{2} \sum_{i,j} 
  \frac{\partial^2}{\partial z_i \, \partial z_j}
  \{ [\ML(\vz,t) \, \MQ \, \ML^\T(\vz,t)]_{ij} \, (\bullet) \}.
\end{equation}
\cref{eq:fpk-op2} allows for various kind of approaches for direct approximation of the FPK equation either by basis function approximations, finite differences, or other methods (Sec.~9.6 in \cite{Sarkka+Solin:2019} provides examples of using point collocation, Ritz--Galerkin, and FEM type of methods for approximating the solution). For example, the results in \cref{fig:teaser} and \cref{fig:FPK-approx} are estimated by a grid discretization over $\vz$ and solving the resulting (finite-dimensional) ODE corresponding to \cref{eq:fpk-op2} by the matrix exponential: $\vp(t) = \exp(\MA (t-t_0))$. See \cref{app:FPK-disc} for details.	%
Even if the widely-used Euler--Maruyama, Milstein, and more general stochastic Runge--Kutta schemes are derived from the It\^o--Taylor series, the resulting methods can still be viewed as an approximation of $p(\vz,t)$.

\subsection{Assumed Density Approximation of FPK}
For the purpose of modelling latent space dynamics of systems of the kind in \cref{fig:workflow}, we note that there the common practice of solving the latent SDE through costly simulation/sampling and then employing a variational (Gaussian) approximation in the encoder/decoder seems contradictory. That is, it might be unnecessary to sample realizations of the trajectory dynamics, if the interest is only in the time-marginals of the process. 
Thus, we seek to directly characterize the first two moments of the solution to the FPK equation in \cref{sec:FPK}. We replace the FPK solution with a Gaussian approximation of form
\begin{equation}
  p(\vz,t) \approx \N(\vz \mid \vm(t),\MP(t)),
\end{equation}
where $\vm(t)$ and $\MP(t)$ are interpreted as a mean and covariance of the state of the solution at time $t$. This kind of approximation is commonly referred to as a Gaussian {\em assumed density} approximation (see, \eg, \cite{Kushner:1967,Sarkka+Sarmavuori:2012}), because the computations are done under the assumption that the state distribution is  Gaussian. 
Assumed density approximations are common in signal processing--driven SDE methodology, and we refer the reader to Sec.~9.1 in \cite{Sarkka+Solin:2019} for a detailed overview. 
Following \cite{Sarkka+Sarmavuori:2012}, we revisit the idea that a Gaussian process approximation to the SDE \cref{eq:SDE} can be obtained by integrating the following differential equations from the initial conditions $\vm(t_0) = \E[\vz(t_0)]$ and $\MP(t_0) = \Cov[\vz(t_0)]$ to the target time $t$:
\begin{align}
  \frac{\diff \vm}{\diff t} &= \int \vf(\vz,t) \, \N(\vz \mid \vm,\MP) \diff \vz \quad \text{and} \label{eq:mean} \\
  \frac{\diff \MP}{\diff t}
   &= \int \vf(\vz,t) \, (\vz - \vm)^\T \, \N(\vz \mid \vm,\MP) \diff \vz \nonumber \\
   & \quad + \int (\vz - \vm) \, \vf^\T(\vz,t) \, \N(\vz \mid \vm,\MP) \diff \vz 
           + \int \ML(\vz,t) \, \MQ \, \ML^\T(\vz,t) \, \N(\vz \mid \vm,\MP) \diff \vz. \label{eq:cov}
\end{align}
These equations for the evolution of the first moments of the solution to the SDE can be interpreted as expectations over the drift and diffusion dynamics of the SDE, and can be derived from the FPK in \cref{eq:fpk}. Conveniently, these expressions are {\em not} stochastic, but instead take the form of an ODE system that---given the integrals are tractable---can be solved with out-of-the-box ODE solvers.
However, even if \cref{eq:mean,eq:cov} provide a generic Gaussian assumed density approximation framework for SDEs, an implementation of the method requires solving the following kind of $d$-dimensional Gaussian integrals:
\begin{equation}\label{eq:gauss-int}
  \E_{\N}[\bullet] 
  = \int [\bullet] \, \N(\vz \mid \vm,\MP) \diff \vz.
\end{equation}
In the following sections we will consider two approaches (local linearization and moment matching with symmetric quadrature) which scale linearly in the number of latent dimensions $d$. 

\subsection{Linearizing the FPK Equation}
\label{sec:lin}
Local linearization around the $\vm$ (via a Taylor series approximation) is a classical approach widely used for this type of Gaussian integrals in machine learning and filtering theory \citep{Jazwinski:1970,Maybeck:1982}. 
If the function $\vf(\vz,t)$ is differentiable, the covariance differential equation can be simplified by using Stein's lemma \cite{Papoulis:1984} such that
\begin{equation}
  \textstyle\int \vf(\vz,t) \, (\vz - \vm)^\T \, \N(\vz \mid \vm,\MP) \diff \vz 
  = \left[ \int \MF_\vz(\vz,t) \, \N(\vz \mid \vm,\MP) \diff \vz \right] \, \MP,
\end{equation}
where $\MF_\vz(\vz,t)$ is the Jacobian of $\vf(\vz,t)$ with respect to $\vz$. Linearizing around the mean $\vm$ and approximating the diffusion as $\ML(\vz,t) \approx \ML(\vm,t)$ gives a linearized form of \cref{eq:mean,eq:cov}:
\begin{equation}
 \frac{\diff \vm}{\diff t} = \vf(\vm,t) \quad \text{and} \quad
 \frac{\diff \MP}{\diff t}
  = \MP \, \MF_\vz^\T(\vm,t) 
   + \MF_\vz(\vm,t) \, \MP 
   + \ML(\vm,t) \, \MQ \, \ML^\T(\vm,t),
\end{equation}
which provides a direct way of propagating the moments of the latent SDE through an ODE for the mean and covariance without the need of drawing multiple sample trajectories. The resulting ODE is $(d+d^2)$-dimensional, and only requires one evaluation of the drift, diffusion, and Jacobian per step.

\subsection{Matching Moments of the FPK Equation}
\label{sec:match}
The local linearization approach given in the preceding section is efficient, but fully {\em local}. An alternative way of constructing an assumed density approximation to $p(\vz,t)$ is to directly match the moments by solving the Gaussian integrals in \cref{eq:mean,eq:cov} by Gaussian quadrature methods. The approximation to \cref{eq:gauss-int} would take the form $\int \vg(\vz,t) \, \N(\vz \mid \vm,\MP) \diff \vz \approx \sum_i w^{(i)} \, \vg(\vz^{(i)},t)$, for an arbitrary integrand $\vg(\vz,t)$, weights $w^{(i)}$, and so called sigma points $\vz^{(i)} = \vm {+} \sqrt{\MP} \, \vxi_i$. Here $\sqrt{\MP}$ denotes a square-root factor of $\MP$ such as the Cholesky decomposition. The multi-dimensional Gaussian quadrature (or {\em cubature}, see \cite{Cools:1997}) rule is characterized by the evaluation points and their associated weights $\{(\vxi_i,w_i)\}$.
We write \cref{eq:mean,eq:cov} in a Gaussian assumed density form which matches the moments by quadrature as follows \cite{Sarkka+Solin:2019}:
\begin{align}
  \textstyle\frac{\diff\vm}{\diff t} &=
     \textstyle\sum_{i} w^{(i)} \, \vf(\vm {+} \sqrt{\MP} \, \vxi_i, t) \quad \text{and} \\
  \textstyle\frac{\diff\MP}{\diff t} &=
     \textstyle\sum_{i} w^{(i)} \, \vf(\vm {+} \sqrt{\MP} \, \vxi_i, t) \,
     \vxi_i^\T \, \sqrt{\MP}^\T \nonumber \\
 &\textstyle+ \sum_{i} w^{(i)} \, 
    \sqrt{\MP} \, \vxi_i \, \vf^\T(\vm {+} \sqrt{\MP} \, \vxi_i, t) 
  + \sum_{i} w^{(i)} \, 
   \ML(\vm {+} \sqrt{\MP} \, \vxi_i,t) \, \MQ \, \ML^\T(\vm {+} \sqrt{\MP} \, \vxi_i,t).
\end{align}
The computational complexity of this approach is highly dependent on the choice of quadrature method. A typical choice in ML applications would be Gauss--Hermite quadrature,  which factorizes over the input dimensions leading to an exponential number ($p^d$) of function evaluations/sigma points in the input dimensionality $d$ for a desired order $p$. In order to guarantee scalability, we employ a symmetric 3\textsuperscript{rd} order cubature rule \citep{Arasaratnam+Haykin:2009} which similarly to Gauss--Hermite ($p=3$) is exact for polynomials up to degree 3. The points are given by scaled unit coordinate vectors $\ve_i$ such that
\begin{equation}
  \vxi_i = 
  \begin{cases} 
    \phantom{-}\sqrt{d}\,\ve_i, \quad \text{for~}i=1,\ldots,d, \\  
    -\sqrt{d}\,\ve_i, \quad \text{for~}i=d+1,\ldots,2d,
  \end{cases}
\end{equation}
and the associated weights are $w_i = \frac{1}{2d}$. This approach provides a direct way of propagating the `true' moments of the latent SDE through an ODE for the mean and covariance and without the need of drawing multiple sample trajectories. The resulting ODE is $(d+d^2)$-dimensional, and requires only $2d$ evaluations of the drift and diffusion per step.

\subsection{Analysis of the Computational Complexity}
\label{sec:complexity}

\setlength{\columnsep}{5pt}
\setlength{\intextsep}{0pt}
\begin{wrapfigure}{r}{0.42\textwidth}
  \centering\scriptsize
  \setlength{\figurewidth}{.38\textwidth}
  \setlength{\figureheight}{.66\figurewidth}
  \pgfplotsset{axis on top,scale only axis,y tick label style={rotate=90}, x tick label style={font=\tiny},y tick label style={font=\tiny},legend columns=2,legend style={nodes={scale=0.8, transform shape}}}
  \pgfplotsset{
    legend image with text/.style={
        legend image code/.code={%
            \node[anchor=center] at (0.3cm,0cm) {#1};
        }
    },
  }
  \begin{minipage}[t]{0.42\textwidth}
    \raggedleft
    \resizebox{\textwidth}{!}{
\begin{tikzpicture}

\definecolor{color0}{rgb}{1,0.498039215686275,0.0549019607843137}
\definecolor{color1}{rgb}{0.12156862745098,0.466666666666667,0.705882352941177}

\begin{axis}[
height=\figureheight,
legend cell align={left},
legend style={
  fill opacity=0.8,
  draw opacity=1,
  text opacity=1,
  at={(0.97,0.03)},
  anchor=south east,
  draw=white!80!black
},
log basis y={10},
tick align=outside,
tick pos=left,
width=\figurewidth,
x grid style={white!69.0196078431373!black},
xlabel={Latent dimensionality, \(\displaystyle d\)},
xmin=-14.5, xmax=524.5,
xtick style={color=black},
y grid style={white!69.0196078431373!black},
ylabel={Wall-clock time (s)},
ymin=0.3, ymax=150,
ymode=log,
ytick style={color=black}
]

\addlegendimage{legend image with text=GPU}
\addlegendentry{}
\addlegendimage{legend image with text=CPU}
\addlegendentry{}

\addplot [thick, black, mark=*, mark size=1, mark options={solid}]
table {%
10 8.89
50 9.42
200 11.59
300 12.78
400 14.59
500 16.27
};
\addlegendentry{ }
\addplot [semithick, black, dashed, mark=*, mark size=1, mark options={solid}]
table {%
10 8.61
50 27.81
200 228.94
300 228.94
400 228.94
500 228.94
};
\addlegendentry{Euler--Maruyama}
\addplot [thick, color0, mark=*, mark size=1, mark options={solid}]
table {%
10 1.91
50 2.32
200 7.91
300 20.56
400 51.9
500 100.28
};
\addlegendentry{ }
\addplot [semithick, color0, dashed, mark=*, mark size=1, mark options={solid}]
table {%
10 0.38
50 4.63
200 600
300 600
400 600
500 600
};
\addlegendentry{Moment matching}
\addplot [thick, color1, mark=*, mark size=1, mark options={solid}]
table {%
10 0.33
50 0.68
200 2.05
300 2.92
400 3.89
500 4.84
};
\addlegendentry{ }
\addplot [semithick, color1, dashed, mark=*, mark size=1, mark options={solid}]
table {%
10 0.29
50 0.65
200 2.46
300 5.2
400 8.71
500 14.17
};
\addlegendentry{Linearization}
\end{axis}

\end{tikzpicture}}
  \end{minipage}\\[1em]
  \begin{minipage}[t]{0.42\textwidth}
    \raggedleft
    \resizebox{\textwidth}{!}{
\begin{tikzpicture}

\definecolor{color0}{rgb}{0.12156862745098,0.466666666666667,0.705882352941177}
\definecolor{color1}{rgb}{1,0.498039215686275,0.0549019607843137}

\begin{axis}[
name=fig,
height=\figureheight,
tick align=outside,
tick pos=left,
width=\figurewidth,
x grid style={white!69.0196078431373!black},
xlabel={Number of Euler--Maruyama trajectories, \(\displaystyle n\)},
xmin=-9.5, xmax=419.5,
xtick style={color=black},
y grid style={white!69.0196078431373!black},
ylabel={Wall-clock time (s)},
ymin=1.463, ymax=14.377,
ytick style={color=black}
]
\addplot [thick, color0, mark=*, mark size=1, mark options={solid}]
table {%
10 2.05
50 2.05
100 2.05
150 2.05
200 2.05
250 2.05
300 2.05
350 2.05
400 2.05
};
\addplot [thick, color1, mark=*, mark size=1, mark options={solid}]
table {%
10 7.91
50 7.91
100 7.91
150 7.91
200 7.91
250 7.91
300 7.91
350 7.91
400 7.91
};
\addplot [thick, black, mark=*, mark size=1, mark options={solid}]
table {%
10 8.66
50 9.18
100 9.76
150 10.5
200 11.13
250 11.68
300 12.55
350 13.22
400 13.79
};
\addplot [semithick, black, dotted]
table {%
225 1.463
225 14.377
};
\end{axis}

\node[fill=white,inner sep=0pt,below right=0.5em] at (fig.north west) {Results for $d=200$.};

\node[fill=white,inner sep=0pt,below right=0.5em,yshift=-8em, xshift=12em,text width=10em,scale=0.8] at (fig.north west) {The dotted line shows when moment matching performs equally well as Euler--Maruyama};

\end{tikzpicture}}
  \end{minipage}
  \captionsetup{justification=centering}
  \caption{Empirical timing experiments \\ with error of final margins matched.}
  \label{fig:timings}
  \vspace*{1em}
\end{wrapfigure}
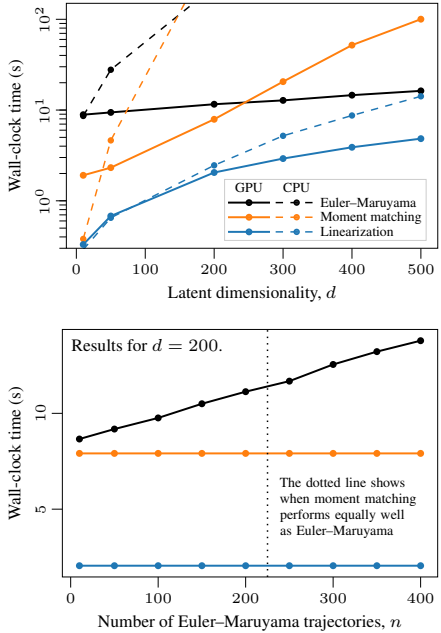%
In terms of the asymptotic computational complexity, the linearization approach in \cref{sec:lin} requires $\bigO(1)$ evaluations of the drift, diffusion, and Jacobian per step. The moment matching approach in \cref{sec:match} requires $\bigO(d)$ evaluations of the drift and diffusion as well as an $\bigO(d^3)$ Cholesky decomposition per step. The simplest Monte Carlo simulation method with $p$ samples requires $\bigO(p)$ evaluations of the drift and diffusion per step. Additionally, the na\"ive requirement for $p$ grows exponentially in $d$. On the other hand, the simulation approach is fully parallelizable over $p$, while the moment matching approach the number of nonparallelizable operations is $\bigO(d^2)$ with the Cholesky decomposition being the bottleneck, and the linearization approach is nonparallelizable. While the linearization approach has the lowest number of function evaluations with respect to $d$, the cost of computing the Jacobian can be prohibitively large for arbitrarily complex models. Nevertheless, the Jacobian is available in closed-form for GPs and may be evaluated reasonably fast for neural network based drifts, see \cref{app:jactiming} for empirical computational costs of evaluating the Jacobian when growing the network size.
Thus we expect the FPK approximation schemes to always be beneficial in CPU-only cases (incl.\ CPU multi-threading and embedded devices). In multicore GPU use, for low-dimensional $d$, sampling remains appealing if GPU memory does not become a bottleneck. %

\section{Experiments}
\label{sec:experiments}
The goals of the experiments are three-fold: We first provide a study of the computational complexity. Then, we look into properties of the GP-SDE model from \cref{sec:gp-sde}, where the experiments are concerned with showcasing model specification rather than inference. Finally, we consider two benchmark problems with high-dimensional inputs for learning a latent SDE model, where we test the performance of the approximations presented when the model is not defined by GPs, as the SDE methods presented in \cref{sec:methods} are model-agnostic.

\paragraph{Timing Experiments}
To confirm the analysis in \cref{sec:complexity} and provide a practical insight, we run numerical experiments with the error of final marginal mean/covariance controlled. We use  a high-dimensional model of $d$ independent Bene\v{s} SDEs ($\!\dd z(t) = \tanh(z)\dd t + \dd\beta(t)$, see \cite{Sarkka+Solin:2019}) with different $z_0$ per dimension. The model is non-linear and solution-space multi-modal, but both $p(\vz,t)$ and the marginal moments ($\vm(t), \MP(t)$) are available in closed form (see \cref{app:timing}). 
In comparison to Euler--Maruyama, we control the number of trajectories $n$ by bounding the error (in terms of KL divergence) between Euler--Maruyama and the ground-truth to match the error in our moment matching approach, and consider the methods equivalent in terms of the quality of the solution. See \cref{app:timing} for the required number of trajectories to match the KL divergence for each dimensionality plotted.
\cref{fig:timings} shows GPU/CPU wall-clock times (GPU: NVIDIA Tesla~V100 32~GB with Intel Xeon Gold 6134 3.2~GHz; CPU: Xeon Gold 6248 2.50GHz).
We implement the models in PyTorch~\cite{NEURIPS2019_9015} and report means of 10 repetitions (std negligible). In low-dimensional cases, both approximation methods outperform sampling, whereas in high dimensions GPU parallelisation becomes dominant and only the linearized approximation remains highly competitive. This example should favour sampling: In the Bene\v{s} model, the number of trajectories, $n$, per $d$ in Euler--Maruyama remains low ($n$ is linear in $d$), which is due to the diagonal (independent) diffusion matrix. In a correlated latent space $n$ would grow super-linearly (even exponentially), which would further push the difference between methods.

\paragraph{GP-SDE Model Specification}
We consider an GP-SDE model with just 8 observations of the dynamics, where the lack of data can be compensated for with encoding prior knowledge into the model. We use the model formulation given in \cref{sec:gp-sde}, and study the effect of GP priors, the first of which is an independent squared exponential (RBF) prior for each dimension which encode continuity and smoothness in the velocity field. The second GP prior is the multi-dimensional curl-free kernel \citep{Wahlstrom:2015} (see \cref{app:gp-sde-experiments})
which encodes the assumption of a curl-free random vector field. This property can be interpreted as `loop aversion' in the GP-SDE context. The third prior, is a multi-dimensional divergence-free kernel \citep{Wahlstrom:2015}
which encodes the assumption of no divergence in the random vector field. This property can be interpreted as `energy preservation' or source-freeness. These properties are visible in \cref{fig:effect-of-prior}, where the hyperparameters are fixed to same values for all models.%

\paragraph{Assumed Density Approximation of the FPK}
We provide an illustrative example of the moment evolution methods in a GP-SDE model with 8 observations along a bean curve and independent squared exponential GP priors per dimension. As a baseline, we solve the FPK in \cref{eq:fpk-op2} by finite-differences discretization in $\vz$ (see \cref{app:FPK-disc} for details).
\cref{fig:FPK-approx} shows the evolution of the SDE solution over the time-course of 5~s. The probability mass dissolves quickly, which is hard to interpret from the top-row figure alone. Comparison between the point clouds and the top row shows that even with 1000 trajectories and just a two-dimensional space, it is hard to capture detailed structure in the SDE solution. The bottom row compares the local linearization (\cref{sec:lin}) and the moment matching (\cref{sec:match}) assumed density approximations to the exact Gaussian approximation of the FPK solution. The linearized approach is mode-seeking (matches local curvature), while the moment matching approach captures the overall structure of the optimal Gaussian approximation.

\begin{figure*}[!t]
  \tiny
  \begin{subfigure}[b]{.25\textwidth}
  	\centering
	\includegraphics[height=\textwidth]{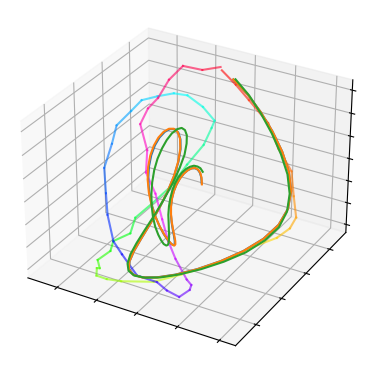}
	\vspace*{\fill}
    \caption{Latent space trajectories}
    \label{fig:rotating-mnist-a}
  \end{subfigure}
  \hfill
  \begin{subfigure}[b]{.75\textwidth}
    \centering
    \setlength{\figurewidth}{.055\textwidth}
    \setlength{\figureheight}{\figurewidth}
    \begin{tikzpicture}[outer sep=0]
       \tikzstyle{box} = [inner sep=0pt, minimum width=\figurewidth,minimum height=\figureheight]

       \foreach \y [count=\j] in {sde,cubature,linearized} {

         \node[box] at ({-.75\figurewidth},{-2.2*\figureheight*\j}) { \includegraphics[width=.95\figurewidth]{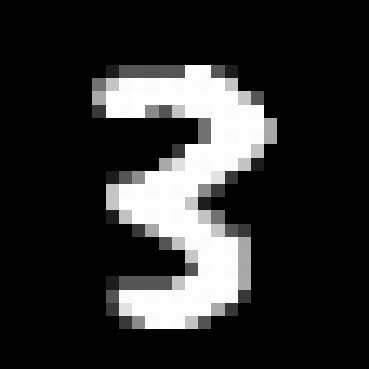} };           

         \foreach \x [count=\i] in {4,8,12,16,20,24,28,32,36,40,44,48,52,56,60,64} {  
           
           \node[box] at ({\figurewidth*\i},{-2.2*\figureheight*\j}) {\includegraphics[width=.95\figurewidth]{fig/mnist/first-\y-mean-\x}};
           \node[box] at ({\figurewidth*\i},{-2.2*\figureheight*\j-\figureheight}) { \includegraphics[width=.95\figurewidth]{fig/mnist/first-\y-std-\x} };
           
         }
       }

       \node at (-.75\figurewidth,-1.5\figureheight) {Input};
       \node at (1.5\figurewidth,-1.5\figureheight) {Predictions $\rightarrow$};   

       \tikzstyle{vlabel} = [rotate=90,text width=2\figureheight,align=center]
       \node[vlabel] at (.15\figurewidth,{-2.2\figureheight*1-.5\figureheight}) {Baseline (1000$\times$~EM)};
       \node[vlabel] at (.15\figurewidth,{-2.2\figureheight*2-.5\figureheight}) {Moment matching};
       \node[vlabel] at (.15\figurewidth,{-2.2\figureheight*3-.5\figureheight}) {Linearization};
    \end{tikzpicture}\\
    \caption{Forward prediction}%
    \label{fig:rotating-mnist-b}
  \end{subfigure}
  \vspace*{-2em}
  \caption{Results on rotating MNIST. (a)~shows the latent space prediction mean trajectories for one test image. Evolution of the true trajectory is shown in HSV colour  and the predicted trajectory by Euler--Maruyama, moment matching, and linearization scheme is shown in green, blue, and orange (all  overlapping one another). (b)~shows the progression of the prediction (mean and std dev) of the test image, when it traverses the learned dynamics in the latent space. Both the moment matching and linearization schemes match the baseline (with 1000 Euler--Maruyama trajectories).}
  \label{fig:rotating-mnist}
\end{figure*}

\paragraph{Rotating MNIST}
In the spirit of \cref{fig:workflow}, we run the proposed methods on Rotating MNIST (\cite{MNIST}, available under CC~BY-SA~3.0), similar to \cite{yildiz2019ode2vae,casale2018gaussian}. The data set consists of various handwritten digit `3's rotated uniformly in 64 angles.
We train a VAE~\cite{kingma2014autoencoding} first by freezing the latent space dynamics, allowing us to generate the latent samples for learning the dynamics by applying the trained VAE. Then freezing the VAE encoder/decoder and training a $16$-dimensional GP-SDE model in the latent space with independent squared exponential GP priors (see \cref{app:mnist}).
In \cref{fig:rotating-mnist} we feed in one observation and let it follow the learned dynamics of rotation. As the baseline, we use $1000$ trajectories computed using Euler--Maruyama. \cref{fig:rotating-mnist-a} demonstrates the model's capability to learn the latent trajectory, and we show the trajectories for all the methods in three latent dimensions with most variation. The trajectories for the three methods overlap, and qualitatively the results are identical in \cref{fig:rotating-mnist-b}.
Quantitatively the brute-force sampling baseline gives slightly better MSEs over images and final-step mean negative log predictive densities (NLPD, see \cref{tbl:mnist}).

\paragraph{Motion Capture Data}
The CMU walking data set (\cite{CMUmocap}, CMU MoCap available under CC BY-ND~4.0) is a real-world noisy data set with $50$ sensors that track a human subject's walking. As in \citet{yildiz2019ode2vae} and \citet{li2020sdeadjoint}, we model the sequences of a single subject, $35$, for which there are $16$ train set, three validation set and four test set sequences. The task is to predict the state of the system in the future given three initial points. 
In this experiment, we demonstrate that replacing SDE solver--based methods by an assumed density approximation, a latent neural SDE system can be learned efficiently without sampling trajectories. For this purpose, and for better comparability to earlier work, the latent SDE drift and diffusion are neural networks.
As in \citet{li2020sdeadjoint}, we regularize the learned posterior process by a prior process. The loss function consists of three terms: reconstruction loss, VAE encoded initial position KL-divergence, and the KL-divergence between posterior/prior processes.  The moments of the posterior SDE approximation are denoted by  $\vm(t)$, $\MP(t)$, those of the prior process $\vm_{*}(t)$, $\MP_{*}(t)$, and the observation times by $\{ t_j\}_{j=0}^{m}$. The loss becomes
\begin{multline}
\mathcal{L} = -\textstyle\sum_{j=0}^{m} \log p(\vx(t_j) \mid \vz(t_j))  
		+ \KL{q(\vz(t_0)\mid \vx(t_0))}{p(\vz)}  	\\
		 + \textstyle\sum_{j=1}^m\gamma \, \KL{\N(\vm(t_j), \MP(t_j))}{\N(\vm_{*}(t_j), \MP_{*}(t_j))},
\end{multline}
where $\vz(t_j)$ and $\vx(t_j)$ are the latent codes and observations, respectively, $q$ denotes the conditional encoder distribution, $p(z)$ is the normal distribution, and $ \log p(\vx(t_j) \mid \vz(t_j)) $ is the model likelihood of the observations, given latent codes.
While training, the parameters to optimize include those of a latent neural network which defines the prior and posterior dynamics, and of the VAE which encodes an initial point and decodes at each discrete time step corresponding to a train set frame. As the VAE is trained simultaneously to the latent dynamics, the approach does not provide a set of true latent samples to compare to, in contrast to the MNIST experiments.
The VAE encoder prior is the normal distribution, whereas the encoder posterior is acting as the initial distribution for the latent SDEs. For the encoder design, we use a fully-connected neural network, which encodes the three first data points to create latent state and context vectors. As the prior process was a SDE with zero drift and $\sigma \MI$ diffusion. The context is passed through the dynamics, similar to the treatment of velocity in \citet{yildiz2019ode2vae} (see \cref{app:mocapappendix}).

The result in \cref{tbl:mocap-results} is competitive considering that solving the latent SDE with the linearization approach roughly matches the required computation budget for {\em one} stochastic Runge--Kutta sample in the other methods. This is highlighted in \cref{tbl:wall-clock} which shows the wall-clock times for the model used in this experiment both in a CPU and a GPU setting (see hardware description in previous experiment).

\begin{table}[t]
  \begin{minipage}[t][4.75cm][t]{.5\textwidth}
  \centering\scriptsize  
  \caption{Rotating MNIST results.}  
  \label{tbl:mnist}
  \begin{tabularx}{\textwidth}{l c c} 
			\toprule
			\sc Inference scheme &  \sc MSE & \sc NLPD ($t=64$) \\  
			\midrule
			Euler--Maruyama & $0.046 \pm 0.006$ & $33.0 \pm 5.4$\\
			Moment matching & $0.051 \pm 0.007$ & $52.7 \pm 9.5$ \\
			Linearization   & $0.052 \pm 0.007$ & $54.5 \pm 9.9$ \\
			\bottomrule 
  \end{tabularx}
  \vspace*{\fill}
  \caption{Wall-clock timings for MOCAP.}
  \label{tbl:wall-clock}
  \setlength{\tabcolsep}{2pt}    
  \begin{tabularx}{\textwidth}{l c c | c c c} 
  \toprule
  \multicolumn{3}{l}{\sc Time/s${\pm}$std} & \multicolumn{3}{c}{\sc Number of E-M paths} \\ 
   & \sc Lin. & \sc Mom.\ mat. & 1 & 25 & 200 \\  
  \midrule
  GPU &   $2.1{\pm}.1$ &  $6.0{\pm}.1$ & $37.1{\pm}.1$        & $39.9{\pm}.1$ & $ 40.4{\pm}.7\phantom{1}$ \\
  CPU &   $1.8{\pm}.1$ &$1.8 {\pm}.1$& $27.7{\pm}.7$ & $\phantom{.2}38.6{\pm}1.5\phantom{.}$ & $94.2 {\pm}3.5 $ \\
  \bottomrule
  \end{tabularx}

  \end{minipage}
  \hfill
  \begin{minipage}[t][4.75cm][t]{.45\textwidth}
  \caption{Test MSE on 297 future MOCAP points averaged over $50$ samples. $95\%$ confidence interval reported based on t-statistic. $^\dagger$results from~\cite{yildiz2019ode2vae}, $^\ddagger$results from~\cite{li2020sdeadjoint}}
  \label{tbl:mocap-results}
  \vspace*{\fill}
  \centering\scriptsize
  \renewcommand{\arraystretch}{.75}  
  \setlength{\tabcolsep}{6pt}  
  \begin{tabularx}{\textwidth}{l l} 
  \toprule
  \sc Method & \sc Test MSE \\
  \midrule
  DTSBN-S~\cite{gan2015deep} & $34.86 \pm 0.02^\dagger$  \\
  {np}ODE~\cite{heinonen2018learning} & $22.96^\dagger$ \\
  {Neural}ODE~\cite{chen2018neuralode} &  $22.49 \pm 0.88^\dagger$ \\
  \ODEtwoVAE~\cite{yildiz2019ode2vae} & $10.06 \pm 1.4^\dagger$ \\
  \ODEtwoVAE-KL~\cite{yildiz2019ode2vae} & $8.09 \pm 1.95^\dagger$\\
  Latent ODE~\cite{rubanovalatentode2019} & $5.98 \pm 0.28^\ddagger$ \\
  Latent SDE~\cite{li2020sdeadjoint} & ${4.03 \pm 0.20}^\ddagger$ \\
  Latent SDE (assumed density)  &  $7.55 \pm 0.05 $ $\phantom{0^\ddagger}$ \\
  \bottomrule
  \end{tabularx}
   
  \end{minipage}
\end{table}

\section{Discussion and Conclusions}
\label{sec:discussion}
In this paper our interest has been in both SDE model specification and approximative inference. We considered GP-SDE models for data-scarce applications that need injection of prior knowledge such as in \cref{fig:effect-of-prior}. For inference, we built upon the established methodology of assumed density approximations in signal processing for directly approximating the solution distribution of an It\^o SDE, where the methods apply to any latent space SDE models.

We put interest in {\em weak} solution concepts for SDEs. We considered both linearization and moment matching based methods for capturing the first two moments of the SDE solutions, which respectively require only $\mathcal{O}(1)$ and $\mathcal{O}(d)$ evaluations of the model functions per solver step in latent space dimensionality $d$. Furthermore, they only require solving of one ODE rather than simulating multiple trajectories. This makes them orders of magnitudes lighter than the current state-of-the-art in neural SDEs, which we analyzed both through theoretical bounds in \cref{sec:complexity}, ran numerical experiments with the final error controlled (\cref{fig:timings}), and highlighted the practical wall-clock time in a MOCAP experiment. We further argued that a Gaussian assumption makes sense in applications of the form in \cref{fig:workflow} as one is typically employed anyway in the encoder--decoder.

There are some key differences between our assumed density approach and commonly used sampling approaches to solving neural SDE models. While stochastic Runge--Kutta methods are typically either concerned with strong (pathwise) or weak (in distribution) solutions, we leverage the even weaker solution concept of only tracking the first two moments of the solution. This simplification of turning solving the SDE into a deterministic ODE problem of its moments, comes with some remarkable computational savings, and often suffice in practical modelling. 
Then again, the proposed method does not lend itself well to cases where pathwise sample trajectories are required.

We recognize that we cannot give guarantees for the approximated Gaussian integrals to capture the evolution of the true moments outside particular special cases (\eg,  3\textsuperscript{rd} order cubature is exact for polynomials up to order 3). Yet the performance on practical applications is considered reliable, and these kinds of approached are commonly employed across assumed density filtering in signal processing. The use of the approximations schemes we present inherently makes the assumption that the time-marginals of the process are of higher interest compared to the pathwise solutions to the SDE. While the sampling-based methods are less efficient than assumed density approximations, their use is well-justified in applications where sampling trajectories is the purpose of the application.

Codes for the methods and experiments in this paper are available at
\url{http://github.com/AaltoML/scalable-inference-in-SDEs}.

\begin{ack}
Authors acknowledge funding from Academy of Finland (grant numbers 324345 and 339730). We also wish to thank the anonymous reviewers for their comments on our manuscript, and \c{C}a\u{g}atay Y{\i}ld{\i}z and Xuechen Li for providing useful details on the MOCAP experiment. We acknowledge the computational resources provided by the Aalto Science-IT project. ET has been employed part-time at Sellforte Oy during the project. The MOCAP data was obtained from mocap.cs.cmu.edu (created with funding from NSF EIA-0196217).

\end{ack}

\phantomsection%
\addcontentsline{toc}{section}{References}
\begingroup
\small
\bibliographystyle{abbrvnat}

\endgroup

\clearpage

\appendix

\nipstitle{
    {\Large Supplementary Material:} \\
    Scalable Inference in SDEs by Direct Matching of the Fokker--Planck--Kolmogorov Equation}
\pagestyle{empty}

This supplementary document is organized as follows. 
\cref{app:details} provides further details and derivations to facilitate understanding of the methodology.
\cref{app:experiments} includes full details on the experiments, baseline methods, data sets, and additional results.

\section{Methodological Details}
\label{app:details}

We provide details in terms of the concept of `solution' to an SDE, how we use a finite-differences approach for solving the FPK as baseline, and comments on the existence properties of the GP-SDE model formulation.

\subsection{On the Concept of a Solution of an SDEs}
\label{app:solutions}
As illustrated in \cref{fig:teaser} in the main paper, the concept of a `solution' to an SDE is broader than that of an ODE. We restrict our interest to It\^o type SDEs, and consider two types of solution concepts here: {\em (i)}~Strong (path-wise) solutions, and {\em (ii)}~weak (in-distribution) solutions.

A strong solution trajectory $\hat{\vz}(t)$ to an SDE resembles an `actual' (ideal, often intractable in practice) solution trajectory. For simulation methods, their order of strong convergence $\gamma$ (see, \eg, p.~132 in \cite{Sarkka+Solin:2019}) can be characterized by looking at the expected error after $M=\nicefrac{1}{\Delta t}$ steps of length $\Delta t$, $\E[|\vz(t_M)-\hat{\vz}(t_M)|] \leq K \Delta t^\gamma$ for some constant $K$. However, it is generally non-trivial to construct high strong order  solution methods to SDEs due to the requirement of solving intractable iterated It\^o integrals in the It\^o--Taylor series expansion. The required step size $\Delta t$ thus remains very small; for example, the Euler--Maruyama method converges with a strong order of $\gamma = \nicefrac{1}{2}$, which makes it tricky to choose a small-enough step size to ensure the trajectories to resemble an actual solution trajectory. In machine learning, we might be interested in path-wise solutions in the case of drawing an example solution trajectory from the method that follows the evolution of a particular realization of the random forces affecting the output. 

However, during training and testing time, we are typically more interested in {\em aggregating} properties over multiple solution trajectories to either capture the {\em typical} behaviour of the model or quantify {\em uncertainties} induced by the random forces in the model. As the model is stochastic, the full solution entails a probability distribution, $p(\vz,t)$, depending on time $t$ and covering the space $\vz$.

The typical approach for characterizing $p(\vz,t)$ in machine learning applications has been through simulation (sampling), where the Euler--Maruyama scheme, the Milstein scheme, or some more general stochastic Runge--Kutta scheme is often used. For simulation methods, the weak order of convergence (see, \eg, p.~137 in \cite{Sarkka+Solin:2019}) can be used for characterizing the method, where it is defined to be the largest exponent $\alpha$ such that $|\E[g(\vz(t_M))] - \E[g(\hat{\vz}(t_M))]| \leq K \Delta t^\alpha$ for any polynomial function $g(\cdot)$. This is a much weaker criterion as it only considers the error in the expectation, and for example, the Euler--Maruyama method converges with weak order convergence $\alpha = 1$. In practice, this means that the moment properties can be captured with a more moderate step-size than the path-wise resemblance of the solutions.

However, if one is only interested in the first moments and/or if the dimensionality of $\vz$ is high, it makes sense to consider an even weaker solution concept, where one is only concerned with the first two moments of $p(\vz,t) \approx \N(\vm(t),\MP(t))$. This is what is done in this paper.

In short, capturing the true pathwise behaviour of an SDE is challenging (NB: simulation schemes do not generally capture this well) and sampling schemes instead generally capture the distribution by a finite set of samples. If you are interested in the first moments only, it can be generally safer to model those directly.

\subsection{Approximating the FPK Solution Through Finite-differences}
\label{app:FPK-disc}

As a baseline in \cref{fig:FPK-approx} we seek to seek direct ways of assessing the behaviour of the solution to the  Fokker--Planck--Kolmogorov PDE or its transition density. For this we could use any tools from the vast literature of partial differential equation approximations. The approach presented here essentially uses finite differences in the input domain of $\vz$ and then solves the resulting homogeneous ODE system directly.

The Fokker--Planck--Kolmorogov equation has the form
\begin{equation}
  \frac{\partial p(\vz,t)}{\partial t} = \mathcal{A}^* p(\vz,t),
\end{equation}
where $\mathcal{A}^*$ is the operator defined in \cref{eq:fpk-op2} in the main paper. We can approximate this equation as a finite-dimensional system, which is a homogeneous linear system. We discretize the state space to a finite grid $\{ (z_1^{(i)},z_2^{(j)}) : i,j=1,2,\ldots,N \}$ and then approximate the derivatives as finite differences. The approximations   can, for example, be given by
\begin{equation}
\begin{split}
    \frac{\partial p(\vz,t)}{\partial z_1} &{\approx} \frac{p(z_1 {+} \Delta z_1,z_2,t) {-} p(z_1 {-} \Delta z_1, z_2, t)}{2\Delta z_1}, \\
  \frac{\partial^2 p(\vz,t)}{\partial z_1^2} &{\approx} \frac{p(z_1 {+} \Delta z_1, z_2,t) {-} 2 p(\vz,t) {+} p(z_1 {-} \Delta z_1,z_2, t)}{\Delta z_1^2},
\end{split}
\end{equation}
and analogously in the other dimension. We can now interpret \cref{eq:fpk-op2} through these finite difference approximations and form a (very) sparse matrix corresponding to the  adjoint operator $\mathcal{A}^*$, where also the drift and diffusion terms are evaluated at the discrete values. Thus the FPK can be rewritten as a linear ODE system
\begin{equation}
  \frac{\diff \vp}{\diff t} = \MA \, \vp,
\end{equation}
where $\MA$ is the finite-difference approximation matrix for the operator $\mathcal{A}^*$. The initial conditions $p(\vz,t_0)$ can also be collected into a vector $\vp(t_0)$. Because our GP-SDE model is time-invariant, the solution to the homogeneous ODE initial value problem is directly given (in closed-form) as
\begin{equation}
  \vp(t) = \exp((t-t_0) \, \MA) \, \vp(t_0).
\end{equation}
To better explain how this works in practice we have added a Jupyter notebook to this supplement which reproduces this approach.

\subsection{On the GP-SDE Model Construction}
\label{app:on-the-construction}
The model which we call a `GP-SDE' model in the main paper has appeared in various forms in literature before. It directly resembles a `random' ODE model, where the random field $\vv_\theta(\cdot)$ has previously typically either been characterized by a Gaussian random field or Gaussian process model (see, \eg, \cite{Ruttor+Batz+Opper:2013,hedge2019deepgp}) or some parametric model (\eg, \cite{li2020sdeadjoint}). Yet, the existence of the corresponding SDE model as in \cref{eq:SDE,eq:drift-diffusion} is non-trivial. 

The GP-SDE model is presented informally in the main paper, and we {\em do not} guarantee the existence of strong unique solutions to the corresponding SDE model. However, under GP increments, the weak solution of this model exists---which can also be directly empirically shown, \eg, by sampling the GP in an Euler fashion vs.\ running Euler--Maruyama on the corresponding SDE. This highlights the {\em practical} aspects of this model, which is probably also why it has been appearing in various previous forms in machine learning literature.

\section{Experiment Details}
\label{app:experiments}

We provide additional details and results for the experiments presented in the paper, and further evaluate the computational costs of linearized approximation. \cref{fig:rotating-mnist-appendix} follows the same structure as \cref{fig:rotating-mnist} in the main paper, just providing further examples from the test set.

\subsection{Empirical Wall-Clock Timing Experiments}
\label{app:timing}
For the timing experiments in \cref{sec:experiments}, we constructed a setup that allowed us to control the approximation error. We used the Bene\v{s} SDEs model (see details on this, \eg, in \cite{Sarkka+Solin:2019}) that has the form 
\begin{equation}
  \dd z(t) = \tanh(z(t))\dd t + \dd\beta(t),
\end{equation}
where $\beta(t)$ is standard Brownian motion and the initial state $z_0$ is known. This model is non-linear in the drift and the solution is not directly apparent. Conveniently, this model has a closed-form solution that we can leverage as a control. The transition density or solution to the FPK equation is given as:
\begin{equation}
   p(z,t) = \frac{1}{\sqrt{2\pi t}} \frac{\cosh(z)}{\cosh(z_0)} \exp\bigg(-\frac{1}{2} t\bigg) \, \exp\bigg(-\frac{1}{2 t}(z-z_0)^2\bigg).
\end{equation}
This solution is bi-modal and thus the moment matching approach will be an approximation to the true solution distribution. Also the first two moments are available in closed-form and given by:
\begin{align}
  \begin{split}\label{eq:benesmoment}
  m(t) &= z_0 + \tanh(z_0) \, t, \\
  P(t) &= z_0^2 + 2 z_0 \tanh(z_0) \, t + t + t^2 - [m(t)]^2.
  \end{split}
\end{align}
This model is one-dimensional in $z$, but we expand it to $\vz \in \R^d$ by considering $d$ independent Bene\v{s} SDE models over $\vz$ with different $z_0^{(d)}$. The initial points of the trajectories were chosen with linear spacing in $[0, 1]$, with a step size of $1/d$. This test setup should be favourable to a stochastic Runge--Kutta approach, where the samples now do not need to account for correlation in the latent space, thus pushing down the required number of sample trajectories, which we expect to be linear in $d$ (the assumption of a diagonal diffusion was encoded in the experiment setup {\it a~priori}, and the independence of the dimensions in the diffusion function was not). We thus claim that this experiment rather highlights the worst case benefits of our method, rather than the best case.

The number of trajectories used in the stochastic Euler--Mauryama was chosen to match the KL divergence of the moment matching approximation. That is, we initially completed the moment matching approximation for a given dimensionality $d$, obtaining the moments $\vm_\textrm{mm}(t)$, $\MP_\textrm{mm}(t)$, which were compared to the closed-form moments in \cref{eq:benesmoment} by the Kullback--Leibler distance. We evaluated the KL divergence at one hundred values of $t$ (with a spacing of $0.1$), and calculated the total divergence as a sum of the divergence at individual points.  

In order to determine the number of trajectories generating moments of a comparable quality to the moment matching approach, we computed the first two moments over a varying number of trajectories. The moments were compared to the true Bene\v{s} SDE moments with the same KL divergence metric as the moment approximation. For each $d$, we matched the number of trajectories $n$ by controlling that
\begin{equation}
  \KL{\N(\vm_\textrm{EM}^{(n)}(t), \MP_\textrm{EM}^{(n)}(t))}{\N(\vm(t), \MP(t))} \leq
  \KL{\N(\vm_\textrm{mm}(t), \MP_\textrm{mm}(t))}{\N(\vm(t), \MP(t))},
\end{equation}
where $(\vm(t), \MP(t))$ are the exact moments, $(\vm_\textrm{EM}^{(n)}(t), \MP_\textrm{EM}^{(n)}(t))$ the moments from the Euler--Maryuama solution with $n$ trajectories, and $(\vm_\textrm{mm}(t), \MP_\textrm{mm}(t))$ the moments from our moment matching approach. As sampling methods are stochastic, we generated $n$ trajectories $10$ times, to account for the uncertainty. For low values of $d$, the standard deviation of the KL divergence was in the range of 10--20\% of the mean value, whereas for higher values of $d$, such as $200$ or $500$, the uncertainty of the metric was reduced to  at most a percentage of the mean.

For this experiment, where the dimensions are uncorrelated, we found that the required number of trajectories to obtain the KL divergence of moment matching approximations was, depending on the dimensionality, between $1d$ and $2d$ (for dimensions [10, 50, 200, 300, 400, 500], the number of trajectories in the same order was [25, 65, 225, 325, 440, 550]). While the linearized approximation generally was less accurate than moment matching, it was equivalent to a nearly as high number of trajectories as moment matching (for example, when $d=200$, linearized approximation was comparable to $210$ trajectories, and moment matching to $225$ trajectories). 

We implement the models in PyTorch~\cite{NEURIPS2019_9015} and report means of 10 repetitions (std over runs negligible and omitted for clarity of presentation). The GPU/CPU wall-clock times that are reported in the main paper were run on a cluster with separate GPU and CPU partitions (GPU: NVIDIA Tesla~V100 32~GB with Intel Xeon Gold 6134 3.2~GHz; CPU: Xeon Gold 6248 2.50~GHz).

\subsection{Additional Jacobian Computation Timing Experiment}
\label{app:jactiming}
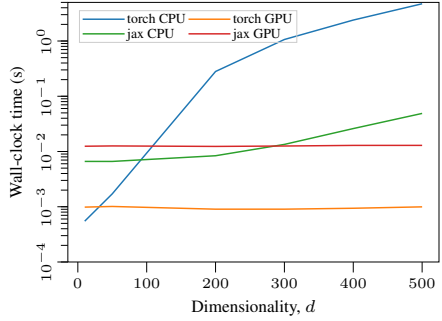
\begin{figure}
  \centering\scriptsize
  \setlength{\figurewidth}{.38\textwidth}
  \setlength{\figureheight}{.7\figurewidth}
  \pgfplotsset{axis on top,scale only axis,y tick label style={rotate=90}, x tick label style={font=\tiny},y tick label style={font=\tiny},legend columns=2,legend style={nodes={scale=0.8, transform shape}}}
  \pgfplotsset{
    legend image with text/.style={
        legend image code/.code={%
            \node[anchor=center] at (0.3cm,0cm) {#1};
        }
    },
  }
  \begin{minipage}[t]{0.42\textwidth}
    \raggedleft
    \resizebox{\textwidth}{!}{
\begin{tikzpicture}

\definecolor{color0}{rgb}{0.12156862745098,0.466666666666667,0.705882352941177}
\definecolor{color1}{rgb}{1,0.498039215686275,0.0549019607843137}
\definecolor{color2}{rgb}{0.172549019607843,0.627450980392157,0.172549019607843}
\definecolor{color3}{rgb}{0.83921568627451,0.152941176470588,0.156862745098039}

\begin{axis}[
height=\figureheight,
legend cell align={left},
legend style={fill opacity=0.8, draw opacity=1, text opacity=1, at={(0.03,0.97)}, anchor=north west, draw=white!80!black},
log basis y={10},
tick align=outside,
tick pos=left,
width=\figurewidth,
x grid style={white!69.0196078431373!black},
xlabel={Dimensionality, \(\displaystyle d\)},
xmin=-14.5, xmax=524.5,
xtick style={color=black},
y grid style={white!69.0196078431373!black},
ylabel={Wall-clock time (s)},
ymin=0.0001, ymax=5,
ymode=log,
ytick style={color=black}
]
\addplot [semithick, color0]
table {%
10 0.000548150804307726
50 0.00171192487080892
200 0.28076344066196
300 1.06537336773343
400 2.39414583312141
500 4.72266554832458
};
\addlegendentry{torch CPU}
\addplot [semithick, color1]
table {%
10 0.000984377331203884
50 0.0010120603773329
200 0.000903632905748155
300 0.000903950797186957
400 0.000939051310221354
500 0.000995424058702257
};
\addlegendentry{torch GPU}
\addplot [semithick, color2]
table {%
10 0.00661603609720866
50 0.00661047299702962
200 0.00838059849209256
300 0.013445774714152
400 0.0260091887580024
500 0.0488907761043972
};
\addlegendentry{jax CPU}
\addplot [semithick, color3]
table {%
10 0.0124544302622477
50 0.0126128726535373
200 0.012360413869222
300 0.0125710699293349
400 0.0128957960340712
500 0.012904511557685
};
\addlegendentry{jax GPU}
\end{axis}

\end{tikzpicture}}
  \end{minipage}
  \caption{Empirical timing experiments of Jacobian evaluations over a single point. The variation between repetitions was negligible, and is omitted for clarity. }
  \label{fig:jactimings}
\end{figure}

As we note in \cref{sec:complexity}, a single step of linearized approximation requires only $\mathcal{O}(1)$ drift, diffusion and Jacobian evaluations. When the drift function is defined by a neural network, the scaling of the computational costs from evaluating the Jacobian is not inherently clear, as the network size is grown. In order to better assess the empirical computational costs of linearized approximation in dimension $d$, we evaluate the Jacobian of a neural network with two hidden layers, each with $3d$ nodes. The experiment results presented in \cref{fig:jactimings} demonstrate that when GPU resources are available, the cost of evaluating the Jacobian in linearized approximation is not a bottleneck for growing the network size or approximating SDEs in high-dimensional spaces. The experiment set-up in terms of hardware used is as in  \cref{app:timing}, and the Jacobian was evaluated 10 times, the first of which was discarded due to initialization overhead.

\subsection{GP-SDE Model Specification}
\label{app:gp-sde-experiments}
This example was included to highlight the idea behind the GP-SDE model in a simple task, where encoding prior knowledge plays a major role in the outcome. We considered a GP-SDE model with just 8 observations of the dynamics, where the lack of data can be compensated for by encoding prior knowledge into the model. We use the model formulation given in \cref{sec:gp-sde}, and study the effect of GP priors, the first of which is an independent squared exponential (RBF) prior for each dimension which encodes continuity and smoothness in the velocity field. The second GP prior is the multi-dimensional curl-free kernel \cite{Wahlstrom:2015}:
\begin{equation}
  \vkappa_\text{cf}(\vz, \vz')=\frac{\sigma^2}{\ell^{2}} e^{-\frac{\|\vz-\vz'\|^{2}}{2 \ell^{2}}}\left[\MI-\left(\frac{\vz-\vz'}{\ell}\right)\left(\frac{\vz-\vz'}{\ell}\right)^\T\right],
\end{equation}
which encodes the assumption of a curl-free random vector field. This property can be interpreted as `loop aversion' in the GP-SDE context. The third prior, is a multi-dimensional divergence-free kernel \citep{Wahlstrom:2015}:
\begin{equation}
  \vkappa_\text{df}(\vz, \vz')=\frac{\sigma^2}{\ell^{2}} e^{-\frac{\|\vz-\vz'\|^{2}}{2 \ell^{2}}} \left[\left(\frac{\vz-\vz'}{\ell}\right)\left(\frac{\vz-\vz'}{\ell}\right)^\T
	+\left((d-1)-\frac{\|\vz-\vz'\|^{2}}{\ell^{2}}\right) \, \MI \right],
\end{equation}
which encodes the assumption of no divergence in the random vector field. This property can be interpreted as `energy preservation' or source-freeness. These properties are visible in \cref{fig:effect-of-prior}, where the hyperparameters are fixed to same values for all models (even if the interpretation differs).

\subsection{Synthetic Race Track}
As an additional example, we run the proposed algorithm on two synthetic race-tracks: oval-shaped and bean-shaped. For both the experiments, we have a true race track and a set of noisy observed race tracks. The latent state, $\vz \in \R^2$, governs the dynamics in the original space. Further, using the noisy observed race tracks, we create a set of observation vectors on which the GP is conditioned. For this, we use the Gaussian process regression model in the GPflow~\cite{GPflow2017} framework with the squared-exponential (RBF) kernel. The hyperparameters of the model are optimized with the {Adam} optimizer with a learning rate of $0.001$, and the kernel hyperparameters are initialized with default values, length-scale and variance 1.0.

\begin{figure*}[!t]
	\centering
	\tikzstyle{box} = [draw=none,rounded corners=1pt,inner sep=1pt]
	\setlength{\figurewidth}{.33\textwidth}
	\setlength{\figureheight}{.8\figurewidth}
	\begin{tikzpicture}[outer sep=0]
	\foreach \x [count=\i] in {observation,obs-vectors,gp-predfield,gp-trajectories,Euler-maruyama,gauss-hermite} {  	
		\node[box] at ({\figurewidth*Mod(\i-1,3)},{\figureheight*int(-(\i-1)/3)}) {\includegraphics[width=.95\figurewidth]{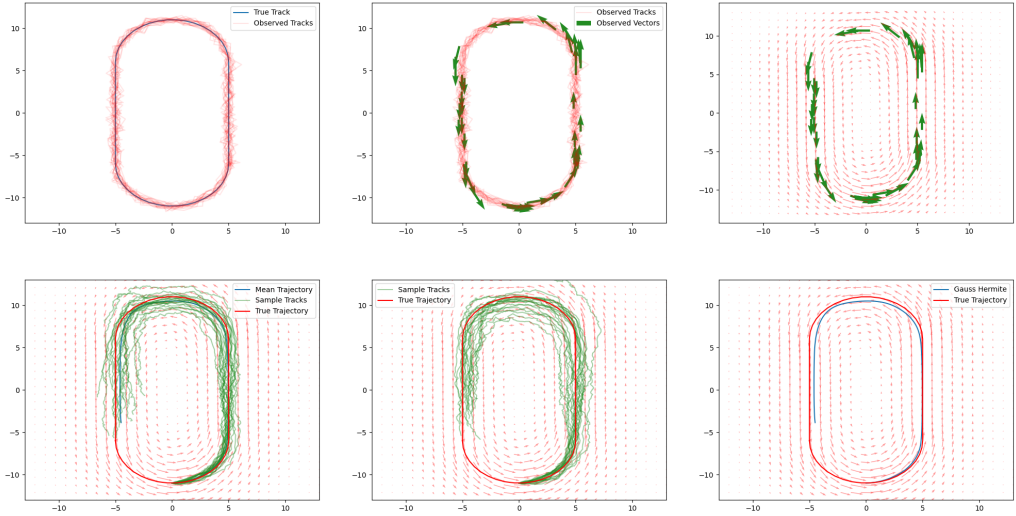}};
	}
	\end{tikzpicture}\\
	\caption{The output of oval shaped synthetic car race track. (a)~True track and noisy observation tracks. (b)~Observation vectors and observation tracks. (c)~Mean predicted vector field of the GP-SDE. (d)~Trajectories predicted using the Euler scheme for the `random ODE' interpretation, with dynamics drawn from GP samples. (e) Trajectories predicted by the Euler--Maruyama scheme for the GP-SDE model. (f)~Mean trajectory path predicted using Gauss--Hermite quadrature by assumed density.}
	\label{fig:OvalCarTrack}
\end{figure*}

\begin{figure*}[!t]
	\centering
	\setlength{\figurewidth}{.33\textwidth}
	\setlength{\figureheight}{.8\figurewidth}
	\tikzstyle{box} = [draw=none,rounded corners=1pt,inner sep=1pt]
	\begin{tikzpicture}[outer sep=0]
	\foreach \x [count=\i] in {observation,obs-vectors,gp-predfield,gp-trajectories,Euler-maruyama,gauss-hermite} {  	
		\node[box] at ({\figurewidth*Mod(\i-1,3)},{\figureheight*int(-(\i-1)/3)}) {\includegraphics[width=.95\figurewidth]{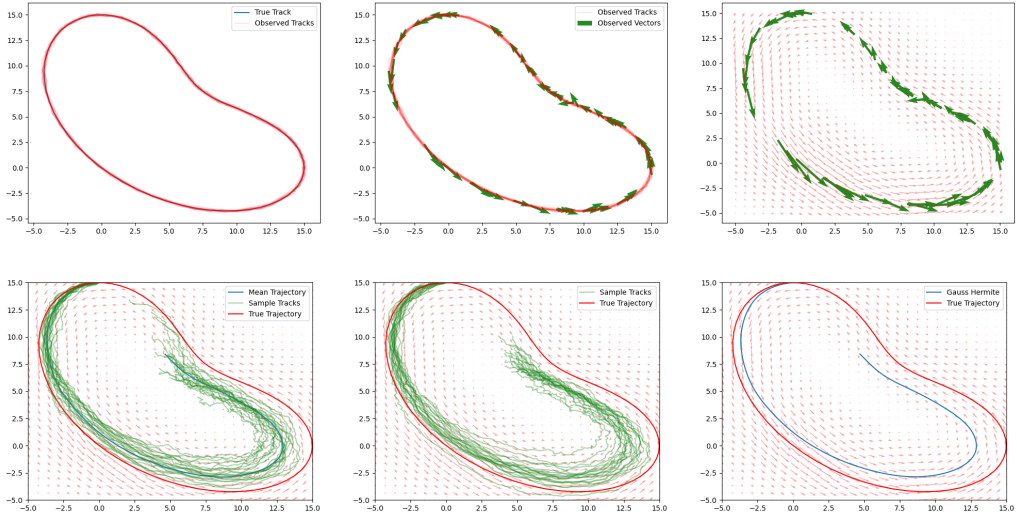}};
	}
	\end{tikzpicture}\\
	\caption{The output of bean shaped synthetic car race track. (a)~True track and noisy observation tracks. (b)~Observation vectors and observation tracks. (c)~Mean predicted vector field of the GP-SDE. (d)~Trajectories predicted using the Euler scheme for the `random ODE' interpretation, with dynamics drawn from GP samples. (e) Trajectories predicted by the Euler--Maruyama scheme for the GP-SDE model. (f)~Mean trajectory path predicted using Gauss--Hermite quadrature by assumed density.}
	\label{fig:BeanCarTrack}
\end{figure*}

\cref{fig:OvalCarTrack} and \cref{fig:BeanCarTrack} showcase the outputs on the two synthetic race tracks. The figures show the true track and noisy observations of the dynamics along the trajectory. On top-right, the mean predicted vector field of the GP-SDE for these observations. Then we visually compare trajectories predicted using the Euler scheme for the `random ODE' interpretation, where the dynamics are drawn from GP samples (see \cref{eq:ode} in the main paper), with trajectories predicted by the Euler--Maruyama scheme for the GP-SDE model. Finally, we plot the mean trajectory path predicted using 4\textsuperscript{th} order Gauss--Hermite quadrature by an assumed density assumption.


\subsection{Rotating MNIST}
\label{app:mnist}

In the spirit of \cref{fig:workflow}, we run the proposed methods on Rotating MNIST (\cite{MNIST}, available under CC~BY-SA~3.0), similar to \cite{yildiz2019ode2vae,casale2018gaussian}. The data set consists of various handwritten digit `3's rotated uniformly in 64 angles.
The training data set is generated by randomly selecting 180 different versions of digit `3' resulting in the total size of the training data set to be $11{,}520$ images. A separate set of 20 digits are chosen to form a test set.
We train a VAE~\cite{kingma2014autoencoding} first by freezing the latent space dynamics, and then freezing the VAE encoder/decoder and training a $16$-dimensional GP-SDE model in the latent space with independent squared exponential GP priors (see \cref{app:mnist}). We implement the VAE model in PyTorch \cite{NEURIPS2019_9015} and use GPflow~\cite{GPflow2017} for the latent GP model. The latent space is chosen to be $d=16$ dimensional and a sparse variational Gaussian process (SVGP) model \cite{hensman2014scalable} with 1000 trainable inducing points is leveraged to scale the GP training. We use independent squared-exponential prior covariance function for each latent dimension dynamics. 
The models are trained with the Adam optimizer \citep{kingma2014adam} (learning rate $0.001$), and the VAE loss function is a weighted sum of binary cross-entropy and KL-divergence, whereas the GP objective function is the ELBO. The training of the two components is disjoint, and the GP is trained on a fixed latent space given by a trained VAE.

\begin{figure*}[!t]
	\tiny
	\begin{subfigure}[b]{.25\textwidth}
		\centering
		\includegraphics[height=\textwidth]{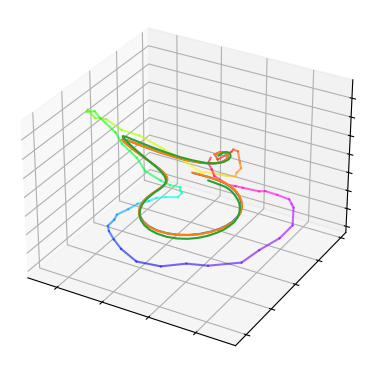}
		\vspace*{\fill}
	\end{subfigure}
	\hfill
	\begin{subfigure}[b]{.75\textwidth}
		\centering
		\setlength{\figurewidth}{.055\textwidth}
		\setlength{\figureheight}{\figurewidth}
		\begin{tikzpicture}[outer sep=0]
		\tikzstyle{box} = [inner sep=0pt, minimum width=\figurewidth,minimum height=\figureheight]

		\foreach \y [count=\j] in {sde,cubature,linearized} {

			\node[box] at ({-.75\figurewidth},{-2.2*\figureheight*\j}) { \includegraphics[width=.95\figurewidth]{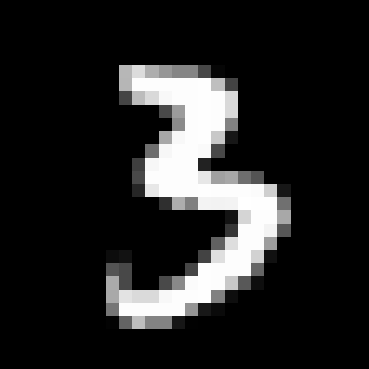} };           

			\foreach \x [count=\i] in {4,8,12,16,20,24,28,32,36,40,44,48,52,56,60,64} {  
				
				\node[box] at ({\figurewidth*\i},{-2.2*\figureheight*\j}) {\includegraphics[width=.95\figurewidth]{fig/mnist/second-\y-mean-\x}};
				\node[box] at ({\figurewidth*\i},{-2.2*\figureheight*\j-\figureheight}) { \includegraphics[width=.95\figurewidth]{fig/mnist/second-\y-std-\x} };
				
			}
		}

		\node at (-.75\figurewidth,-1.5\figureheight) {Input};
		\node at (1.5\figurewidth,-1.5\figureheight) {Predictions $\rightarrow$};   

		\tikzstyle{vlabel} = [rotate=90,text width=2\figureheight,align=center]
		\node[vlabel] at (.15\figurewidth,{-2.2\figureheight*1-.5\figureheight}) {Baseline (1000$\times$~EM)};
		\node[vlabel] at (.15\figurewidth,{-2.2\figureheight*2-.5\figureheight}) {Moment matching};
		\node[vlabel] at (.15\figurewidth,{-2.2\figureheight*3-.5\figureheight}) {Linearization};
		\end{tikzpicture}\\
	\end{subfigure}
	\begin{subfigure}[b]{.25\textwidth}
		\centering
		\includegraphics[height=\textwidth]{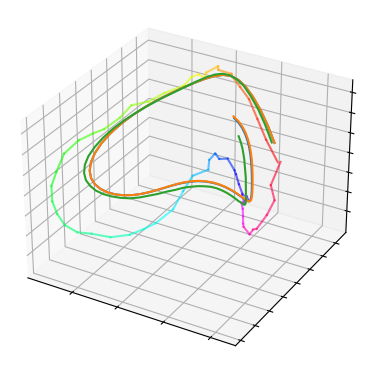}
		\vspace*{\fill}
	\end{subfigure}
	\hfill
	\begin{subfigure}[b]{.75\textwidth}
		\centering
		\setlength{\figurewidth}{.055\textwidth}
		\setlength{\figureheight}{\figurewidth}
		\begin{tikzpicture}[outer sep=0]
		\tikzstyle{box} = [inner sep=0pt, minimum width=\figurewidth,minimum height=\figureheight]

		\foreach \y [count=\j] in {sde,cubature,linearized} {

			\node[box] at ({-.75\figurewidth},{-2.2*\figureheight*\j}) { \includegraphics[width=.95\figurewidth]{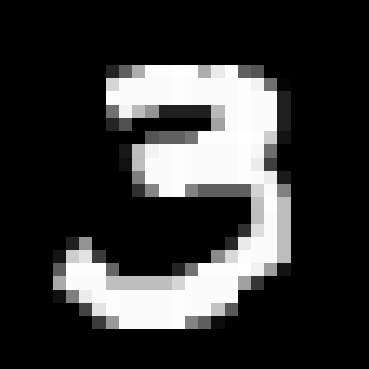} };           

			\foreach \x [count=\i] in {4,8,12,16,20,24,28,32,36,40,44,48,52,56,60,64} {  
				
				\node[box] at ({\figurewidth*\i},{-2.2*\figureheight*\j}) {\includegraphics[width=.95\figurewidth]{fig/mnist/third-\y-mean-\x}};
				\node[box] at ({\figurewidth*\i},{-2.2*\figureheight*\j-\figureheight}) { \includegraphics[width=.95\figurewidth]{fig/mnist/third-\y-std-\x} };
				
			}
		}

		\node at (-.75\figurewidth,-1.5\figureheight) {Input};
		\node at (1.5\figurewidth,-1.5\figureheight) {Predictions $\rightarrow$};   

		\tikzstyle{vlabel} = [rotate=90,text width=2\figureheight,align=center]
		\node[vlabel] at (.15\figurewidth,{-2.2\figureheight*1-.5\figureheight}) {Baseline (1000$\times$~EM)};
		\node[vlabel] at (.15\figurewidth,{-2.2\figureheight*2-.5\figureheight}) {Moment matching};
		\node[vlabel] at (.15\figurewidth,{-2.2\figureheight*3-.5\figureheight}) {Linearization};
		\end{tikzpicture}\\
	\end{subfigure}
	\begin{subfigure}[b]{.25\textwidth}
		\centering
		\includegraphics[height=\textwidth]{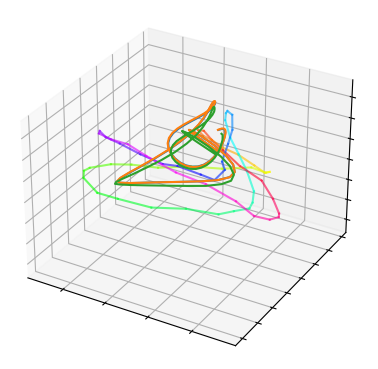}
		\vspace*{\fill}
	\end{subfigure}
	\hfill
	\begin{subfigure}[b]{.75\textwidth}
		\centering
		\setlength{\figurewidth}{.055\textwidth}
		\setlength{\figureheight}{\figurewidth}
		\begin{tikzpicture}[outer sep=0]
		\tikzstyle{box} = [inner sep=0pt, minimum width=\figurewidth,minimum height=\figureheight]

		\foreach \y [count=\j] in {sde,cubature,linearized} {

			\node[box] at ({-.75\figurewidth},{-2.2*\figureheight*\j}) { \includegraphics[width=.95\figurewidth]{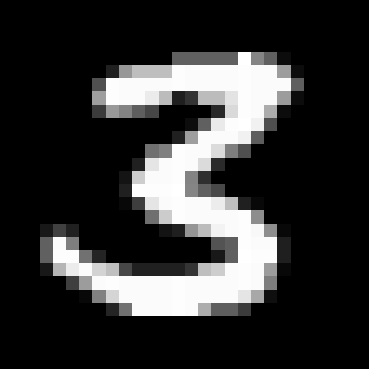} };           

			\foreach \x [count=\i] in {4,8,12,16,20,24,28,32,36,40,44,48,52,56,60,64} {  
				
				\node[box] at ({\figurewidth*\i},{-2.2*\figureheight*\j}) {\includegraphics[width=.95\figurewidth]{fig/mnist/fourth-\y-mean-\x}};
				\node[box] at ({\figurewidth*\i},{-2.2*\figureheight*\j-\figureheight}) { \includegraphics[width=.95\figurewidth]{fig/mnist/fourth-\y-std-\x} };
				
			}
		}

		\node at (-.75\figurewidth,-1.5\figureheight) {Input};
		\node at (1.5\figurewidth,-1.5\figureheight) {Predictions $\rightarrow$};   

		\tikzstyle{vlabel} = [rotate=90,text width=2\figureheight,align=center]
		\node[vlabel] at (.15\figurewidth,{-2.2\figureheight*1-.5\figureheight}) {Baseline (1000$\times$~EM)};
		\node[vlabel] at (.15\figurewidth,{-2.2\figureheight*2-.5\figureheight}) {Moment matching};
		\node[vlabel] at (.15\figurewidth,{-2.2\figureheight*3-.5\figureheight}) {Linearization};
		\end{tikzpicture}\\
	\end{subfigure}
	\caption{Further test set example results on the rotating MNIST data set. The left-hand side shows the prediction trajectories for one test set image in the latent space. Evolution of the true trajectory is shown in HSV colour code and the predicted trajectory by Euler--Maruyama, moment matching, and linearization scheme is shown in green, blue, and orange, respectively (all  overlapping one another). The right-hand figures show the progression of the prediction (mean and marginal std images) of the test set image input, when it traverses the learned dynamics in the latent space. Both the moment matching and linearization schemes match the baseline (computed with 1000 Euler--Maruyama trajectories).}
	\label{fig:rotating-mnist-appendix}	
\end{figure*}

An output of a test point is illustrated in \cref{fig:rotating-mnist}, where we feed in one observation and let it follow the learned dynamics of rotation. As the baseline, we use $1000$ trajectories computed using Euler--Maruyama with step length $0.1$. \cref{fig:rotating-mnist-a} demonstrates the model's capability to learn the latent trajectory, and we show the trajectories for all the methods in three latent dimensions with the most variation. The trajectories for the three methods overlap exactly, which also shows in \cref{fig:rotating-mnist-b} that shows the generated outputs in the observation space together with the associated marginal uncertainties. We include further details in \cref{fig:rotating-mnist-appendix}.

For quantitative comparison, we do a 10-fold cross-validation study on the rotating MNIST data. The full dataset consists of $200$ randomly chosen digits `3' which are split into 10 folds, each fold consisting of $180$ training and $20$ test digits. As discussed, each digit is uniformly rotated around $64$ angles thus making the training dataset size equal to $11{,}520$. Both the models, VAE and latent-GP, are trained independently with the same initial hyperparameter values and an equal number of epochs over different folds. On the test dataset, we perform inference via three schemes: Euler--Maruyama, linearization, and moment matching. Euler--Maruyama acts as a baseline for which $1000$ samples with $0.1$ step-size are generated. For linearization and moment-matching we use the Euler scheme  with $0.1$ step-size for solving the resulting ODE. The MSE values are calculated for the mean. Alongside mean we also characterize the uncertainty estimates by studying the negative log probability density (NLPD) for all the three schemes in the latent space over time. The MSE and (final-step, $t=64$) NLPD results are in \cref{tbl:mnist} in the main paper, where we see that the sampling scheme performs slightly better, especially in terms of NLPD---even if the qualitative results did not show a clear difference.

\subsection{Motion Capture Experiment Details}
\label{app:mocapappendix}
For the motion capture experiments, we used the same pre-processed CMU Walking data set as in \citet{yildiz2019ode2vae}. The relevant hyperparameters and design choices were the weighting of KL-divergence, learning rate of the optimizer, neural network designs, choice of SDE approximation, choice of the ODE solver, and choice of the prior process for regularization. 

\begin{figure*}[!t]
	\centering\footnotesize
	
	\tikzstyle{nn_node} = [rectangle, minimum width=2.5cm, minimum height=1cm, text centered, text width=2.5cm, draw=mycolor0,, fill=mycolor0!10, rounded corners=1pt]

	\tikzstyle{arrow} = [thick,->,>=stealth]
	\begin{tikzpicture}[node distance=1.5cm]
	\node (drift_input) [nn_node] at (0,0) {Latent state, context and time};
	\node (drift_hide1) [nn_node, below of=drift_input] {FC-$30$};
	\node (drift_act1) [nn_node, below of=drift_hide1] {Softmax};
	\node (drift_hide2) [nn_node, below of=drift_act1] {FC-$9$};
	\node (drift_output) [nn_node, below of=drift_hide2] {Drift};

	\draw [arrow] (drift_input) -- (drift_hide1);
	\draw [arrow] (drift_hide1) -- (drift_act1);
	\draw [arrow] (drift_act1) -- (drift_hide2);
	\draw [arrow] (drift_hide2) -- (drift_output);
	
	\node (diff_input) [nn_node, right of=drift_input, xshift=2cm]  {Latent state at index $i$ and time};
	\node (diff_hide1) [nn_node, below of=diff_input] {FC-$30$};
	\node (diff_act1) [nn_node, below of=diff_hide1] {Softmax};
	\node (diff_hide2) [nn_node, below of=diff_act1] {FC-$1$};
	\node (diff_output) [nn_node, below of=diff_hide2] {Diffusion in dimension $i$};	
	
	\draw [arrow] (diff_input) -- (diff_hide1);
	\draw [arrow] (diff_hide1) -- (diff_act1);
	\draw [arrow] (diff_act1) -- (diff_hide2);
	\draw [arrow] (diff_hide2) -- (diff_output);

	\node (encoder_input) [nn_node, right of=diff_input, xshift=2cm] {First three data points: $x_0$, $x_1$, $x_2$};
	\node (encoder_hide1) [nn_node, below of=encoder_input] {FC-$30$};
	\node (encoder_act1) [nn_node, below of=encoder_hide1] {Softmax};
	\node (encoder_hide2) [nn_node, below of=encoder_act1] {FC-$18$};
	\node (encoder_output) [nn_node, below of=encoder_hide2] {Latent state and context mean and log variance};
		
	\draw [arrow] (encoder_input) -- (encoder_hide1);
	\draw [arrow] (encoder_hide1) -- (encoder_act1);
	\draw [arrow] (encoder_act1) -- (encoder_hide2);
	\draw [arrow] (encoder_hide2) -- (encoder_output);

	\node (decoder_input) [nn_node, right of=encoder_input, xshift=2cm] {Latent state};
	\node (decoder_hide1) [nn_node, below of=decoder_input] {FC-$30$};
	\node (decoder_act1) [nn_node, below of=decoder_hide1] {Softmax};
	\node (decoder_hide2) [nn_node, below of=decoder_act1] {FC-$50$};
	\node (decoder_output) [nn_node, below of=decoder_hide2] {x};	
	
	\draw [arrow] (decoder_input) -- (decoder_hide1);
	\draw [arrow] (decoder_hide1) -- (decoder_act1);
	\draw [arrow] (decoder_act1) -- (decoder_hide2);
	\draw [arrow] (decoder_hide2) -- (decoder_output);

	\end{tikzpicture}
	\caption{Neural network designs (from left to right) for the drift, diffusion, encoder and decoder networks. The diffusion network is duplicated $9$ 	times, one for each latent state or context dimension.}
	\label{fig:nn_graph}
\end{figure*}
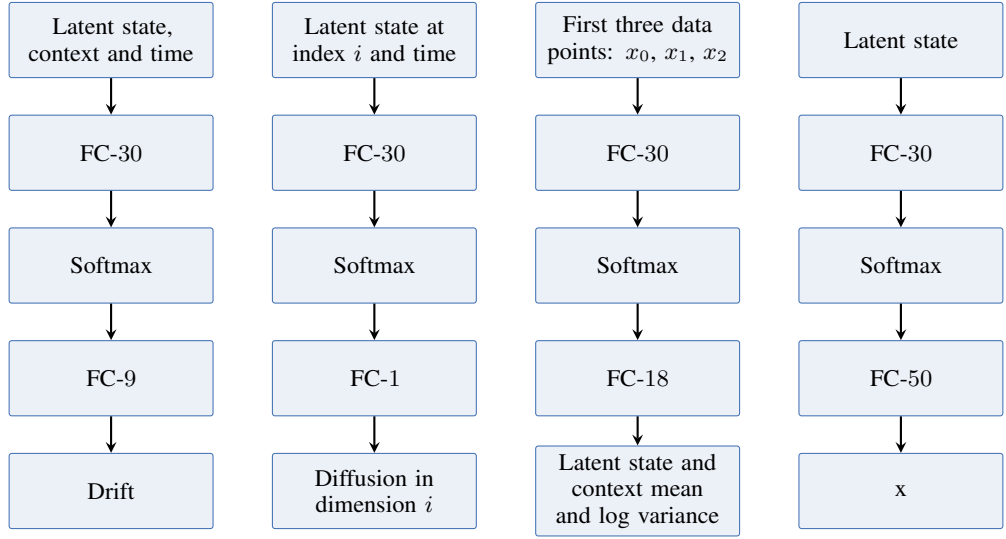

For weighting the KL-divergence, we tested the values $\gamma=\{1.0, 0.1\}$, both with a linear schedule until epoch $200$ and fixed value. The listed MSE was achieved when using $\gamma=1$ fixed and with a learning rate of $0.01$. The drift, diffusion, encoder and decoder networks were all trained simultaneously in $1500$ iterations using the {Adam} optimizer, see \cref{fig:nn_graph} for the detailed network designs. With the exception that we model the change in the latent context, the neural networks are similar to those presented in \citet{li2020sdeadjoint}. 

For the prior process, we used a simple stochastic process with zero drift and $0.1 \MI$ diffusion. In our experiments, we found that using a prior process is fundamental for successful training over an SDE approximation: optimizing solely for maximum likelihood resulted in unrealistic parameter values and lead to numerical instability. As an alternative to zero drift prior processes, we tested a trainable drift network, and inspired by \citet{li2020sdeadjoint}, a prior diffusion network that matches the posterior in state dimensions. While the alternative prior processes produced a more informative latent space, we chose the zero drift prior process both for its simplicity and performance: the lowest MSE was achieved when using a zero drift prior.
The selected ODE solver was a 5\textsuperscript{th} order Runge--Kutta method. 
When running the model implementation for a linearized approximation on TensorFlow and a Tesla V100 GPU, training was completed in approximately 2~hours and 45~minutes. %

We also include a separate timing comparison between the methods under this model, where we control for equal step size and between methods and using the plan Euler/Euler--Maruyama scheme. The results in \cref{tbl:wall-clock} show timings for one pass with a PyTorch implementation and using the same hardware as presented in \cref{app:timing}.

\section{Author Contributions}
The original idea and motivation for this work was conceived by AS, who also wrote a first draft of the paper. ET had the main responsibility of writing the related work section, the computational complexity section, and the MOCAP experiment. PV worked on the GP-SDE models and had the main responsibility of the rotating MNIST experiment. All authors contributed to finalizing the manuscript.

\end{document}